\documentclass[10pt,journal,compsoc]{IEEEtran}
\ifCLASSOPTIONcompsoc
  \usepackage[nocompress]{cite}
\else
  \usepackage{cite}
\fi
\ifCLASSINFOpdf
\else
\fi

\usepackage{titletoc}
\usepackage{longtable} % For tables that span multiple pages
\usepackage{array}     % For thicker hlines/vlines if needed (not used here for booktabs style)
\usepackage{caption}   % For table captions
\usepackage[utf8]{inputenc} % allow utf-8 input
\usepackage[T1]{fontenc}    % use 8-bit T1 fonts
\usepackage{graphicx}       % figures
\usepackage{hyperref}       % hyperlinks
\usepackage{url}            % simple URL typesetting
\usepackage[table]{xcolor}  % colors, \colorbox, and table row colors
\usepackage{colortbl}
\usepackage{wrapfig}
\usepackage{tcolorbox}
\usepackage{algorithm2e}
\usepackage{amsmath,amssymb,amsfonts}
\usepackage{amsthm}
\usepackage{mathtools}
\usepackage{bm}
\usepackage{booktabs}       % professional-quality tables
\usepackage{multirow}
\usepackage{tabularx}       % adaptive-width tables
\usepackage{adjustbox}
\usepackage{enumitem}
\usepackage{rotating}
\usepackage{subfigure}
\usepackage{fontawesome5}   % icons
\usepackage{pifont}
\usepackage{xspace}
\usepackage{titlesec}
\usepackage{nicefrac}       % compact symbols for 1/2, etc.
\usepackage{microtype}      % microtypography
\usepackage[numbers,sort&compress]{natbib}
\definecolor{grey}{rgb}{0.89,0.71,0.57}
\definecolor{pink}{rgb}{1,0.94,1}
\definecolor{purple}{rgb}{0.84,0.78,1}
\definecolor{blue}{RGB}{194,232,247}
\definecolor{white}{rgb}{1,1,1}

\definecolor{mydarkblue}{rgb}{0,0.08,0.45}
\hypersetup{
    colorlinks=true,
    linkcolor=mydarkblue,
    citecolor=mydarkblue,
    filecolor=mydarkblue,
    urlcolor=mydarkblue,
    pdfview=FitH
}
\definecolor{confICLR}{HTML}{D1E8FF} % Light Blue
\definecolor{confNeurIPS}{HTML}{FFD9D9} % Light Red
\definecolor{confICML}{HTML}{D4EFDF} % Light Green
\definecolor{confKDD}{HTML}{FFF3CD} % Light Yellow
\definecolor{confMICCAI}{HTML}{E8DAEF} % Light Purple
\definecolor{confCVPR}{HTML}{D0F0F0} % Light Cyan
\definecolor{confICCV}{HTML}{FFE0F1} % Light Pink
\definecolor{confACMMM}{HTML}{E0E0A0} % Light Olive

\renewcommand{\arraystretch}{1.2}

\definecolor{darkblue}{rgb}{0.0, 0.0, 0.55}
\definecolor{darkred}{rgb}{0.55, 0.0, 0.0}

\definecolor{cpink}{HTML}{FCCDE5}
\definecolor{cred}{HTML}{FFC2BA}
\definecolor{cyellow}{HTML}{FFFFB3}
\definecolor{cblue}{HTML}{B9DEFF}
\definecolor{cgreen}{HTML}{D7F3E7}
\definecolor{cneutral}{HTML}{CFCFCF}
\def\method{DIANA}

\everypar{\looseness=-1}
\titlespacing{\paragraph}{%
  0pt}{%              left margin
  0pt}{% space before (vertical)
  1em}%               space after (horizontal)

\begin{document}
\def\method{CoTIR}
\title{Universal Image Restoration via Internalized Chain-of-Thought Reasoning}

\author{Yu Guo$^\dagger$, Zhengru Fang$^\dagger$, Shengfeng He, \textit{Senior Member, IEEE}, Senkang Hu, \\ Yihang Tao, Phone Lin, \textit{Fellow, IEEE}, Yuguang Fang, \textit{Fellow, IEEE} \\

\thanks{$^\dagger$Yu Guo and Zhengru Fang contributed equally to this work.}
\thanks{Yu Guo, Zhengru Fang, Senkang Hu, Yihang Tao, and Yuguang Fang are with the Hong Kong JC Lab of Smart City and the Department of Computer Science, City University of Hong Kong. Email: \{yu.guo, zhefang4-c, senkang.forest, yihang.tommy\}@my.cityu.edu.hk, my.fang@cityu.edu.hk.}
\thanks{Shengfeng He is with the School of Computing and Information Systems, Singapore Management University. Email: shengfenghe@smu.edu.sg.}
\thanks{Phone Lin is with the Computer Science and Information Engineering, National Taiwan University. Email: plin@csie.ntu.edu.tw.}
}

\markboth{IEEE TRANSACTIONS ON PATTERN ANALYSIS AND MACHINE INTELLIGENCE}
{Shell \MakeLowercase{\textit{et al.}}: Bare Demo of IEEEtran.cls for Computer Society Journals}



\IEEEtitleabstractindextext{%
\begin{abstract}
Image restoration seeks to recover high-quality images from degraded inputs but becomes highly ill-posed under complex, mixed degradations.
While unified all-in-one models are common, their performance declines as degradation complexity increases. 
Recent works adopt Chain-of-Thought (CoT) reasoning for multi-round restoration using specialized modules. 
However, this approach faces two key limitations: \emph{(i) increased computational cost due to multi-step processing}, and \emph{(ii) weak modeling of interactions between degradations during stepwise inference}.
We introduce CoTIR, a universal image restoration framework that internalizes CoT reasoning within a single model.
Concretely, we view image restoration as a specialized subtask of image editing, which implies that a large-scale pre-trained editing model provides a more favorable optimization starting point.
Building on this, we fine-tune the model for restoration and further encode structured CoT-style reasoning into the learning objective via a differentiable formulation inspired by Lagrangian optimization, enabling holistic restoration without chaining specialized restorers.
To facilitate training and evaluation, we further present CoTIR-Bench, a large-scale benchmark comprising 5.2 million samples with CoT-style reasoning traces. 
% Experiments show that CoTIR achieves state-of-the-art performance, robustly disentangling and correcting complex degradations. The source code is available at \url{https://github.com/gy65896/CoTIR}.
Extensive experiments on CoTIR-Bench and broad real composite degradation scenes show that CoTIR achieves stronger perceptual quality and more competitive fidelity than both all-in-one models and multi-round restoration methods. The source code is available at \url{https://github.com/gy65896/CoTIR}.
\end{abstract}

\begin{IEEEkeywords}
Image Restoration, Image Editing, Chain-of-Thought.
%Computer Society, IEEE, IEEEtran, journal, \LaTeX, paper, template.
\end{IEEEkeywords}}

\maketitle

\IEEEdisplaynontitleabstractindextext

% \fi
\IEEEpeerreviewmaketitle

%\MM{When a TPAMI submission is based on a previous conference paper, IEEE requires that the journal paper be a “substantial revision” of the previous publication (30 percent is generally considered “substantial”). TPAMI interprets and applies this requirement on a case-by-case basis with appropriate deference to the author’s viewpoint. Examples of the improvements we expect to see over the conference paper include the following: additional technical details, a clearer explanation of the contribution, more experiments if appropriate, or an updated state-of-the-art. Of course, the authors are also encouraged to make the journal version a significant improvement on the conference paper (for example, by taking the opportunity to integrate their previous work or performing additional substantive work to answer questions that their conference paper raised). Since the journal version is intended to be the definitive, archival version of the research, TPAMI expects that the authors will take this opportunity to further improve their conference paper.}

\section{Introduction}

\begin{figure*}[t]
    \centering
    \includegraphics[width=1\textwidth]{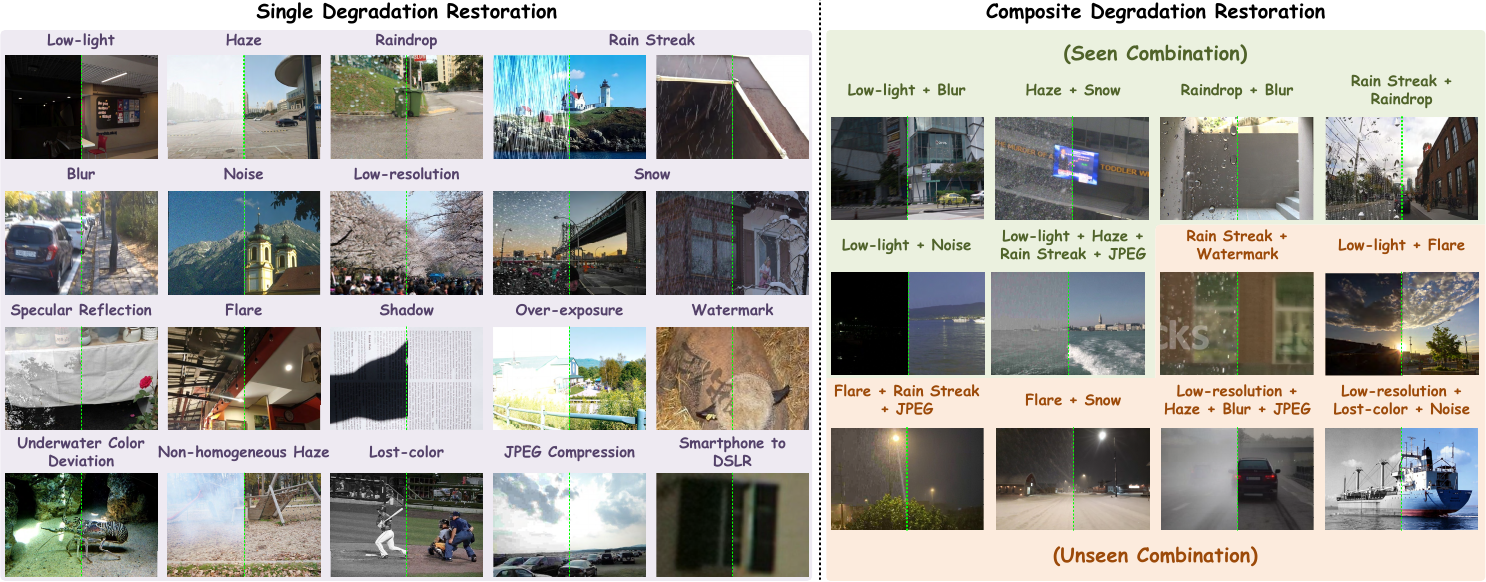}
    \vspace{-6mm}
    \captionof{figure}{We present CoTIR, a universal image restoration framework that internalizes Chain-of-Thought (CoT) reasoning. It not only excels in handling various classic single and composite restoration tasks, but also demonstrates effective generalization to composite degradations with more unseen combinations.}
    \vspace{-6mm}
    \label{fig:visualization}
\end{figure*}

Image restoration aims to reconstruct high-quality images from degraded images. This capability is critical for a broad spectrum of real-world applications, including enhancing consumer photography, improving surveillance clarity, and supporting multi-view perception in autonomous driving \cite{jarvisir2025, Ni_2025_CVPR, Shyam_2024_WACV, fang2026agentcentricobservationadaptationrobust, tao2026learningmutualviewinformation}. In practical scenarios, image degradations arise from a wide range of sources: environmental factors such as rain, haze, and snow; sensor limitations that introduce noise or low contrast; motion-induced blur; and compression artifacts introduced by digital encoding. While considerable progress has been made in addressing isolated degradation types, real-world images are often affected by a complex mixture of multiple corruptions. These degradations are not only co-occurring but also exhibit strong non-linear and spatially varying interactions, making composite image restoration substantially more challenging than handling individual degradations independently. Developing a unified model that is both robust and generalizable to such diverse degradation patterns remains open, demanding further research.

Early efforts \cite{zhu2023learning, kulkarni2022unified, potlapalli2024promptir, guo2024onerestore} attempted to build all-in-one restorers by training a single model across multiple small-scale datasets, covering various degradation types and combinations. Although conceptually appealing, these approaches often suffer from limited generalizability and increased training complexity, because the available data are usually fragmented across tasks and imbalanced in scale. As the number of degradation types and their possible compositions grows, the model is forced to learn disentangled and transferable features from an increasingly entangled data distribution, making optimization more difficult and often leading to compromised robustness on unseen mixtures.

To mitigate this, recent works \cite{zhou2025q, cao2024chain, chen2024restoreagent, zhu2024intelligent} have introduced Chain-of-Thought (CoT) reasoning to conduct multi-round image restoration. Inspired by cognitive problem-solving processes, these methods decompose restoration into a sequence of simpler sub-tasks, handled by specialized modules or tools. This stepwise strategy allows models to focus on one degradation at a time, leading to interpretable and modular pipelines. However, this paradigm introduces two critical limitations. \textit{(i) High computational cost due to multi-step processing}: Each step in the sequence often invokes a separate model or operator, resulting in significant computational overhead during both training and inference. \textit{(ii) Ignoring the interdependence of degradations in stepwise restoration}: Treating each degradation independently overlooks the coupled nature of many real-world corruptions. For instance, deraining may amplify sensor noise, or dehazing may alter image colors in a way that complicates subsequent deblurring. The discrete and sequential formulation prevents coherent modeling of such interactions.

In this paper, we propose \textbf{CoTIR}, a paradigm that reconceptualizes composite image restoration as a structured ``\textbf{Thinking $\rightarrow$ Planning $\rightarrow$ Action}'' process. Rather than explicitly chaining multiple models, CoTIR internalizes this reasoning framework within a single unified system. The model begins by \textit{thinking}, i.e., implicitly disentangling the latent representation into (i) an inherent feature description of the clean content and (ii) a high-level understanding of the degradation patterns. It then performs \textit{planning} by modeling the interplay between these representations to derive a coherent restoration strategy.

This thinking and planning process is operationalized as a set of differentiable, soft constraints, which are incorporated into the training objective via Lagrangian multipliers \cite{boyd2004convex}. These constraints guide the final \textit{action}: an end-to-end restoration capable of addressing both isolated and composite degradations in a single forward pass.
We instantiate CoTIR by fine-tuning the decoder of the powerful \texttt{FLUX} image editing model \cite{batifol2025flux, flux-2-2025}. This choice is based on the hypothesis that \textit{image restoration can be viewed as a specialized case of image editing}, and thus, a pre-trained, instruction-sensitive editing model provides a strong foundation for learning restoration tasks with finetuning. As illustrated in Fig.~\ref{fig:visualization}, CoTIR demonstrates strong adaptability across a wide spectrum of degradation scenarios. It not only handles traditional single-type degradations but also generalizes effectively to complex, previously unseen combinations of real-world corruptions, surpassing existing paradigms in both efficiency and performance. 

In summary, our main contributions are threefold:
\begin{itemize}
    \item We introduce a novel ``Thinking $\rightarrow$ Planning $\rightarrow$ Action'' paradigm that redefines universal image restoration as a structured reasoning problem, enabling holistic, end-to-end modeling within a single, unified framework.
    \item We propose CoTIR, a unified restoration framework that internalizes this reasoning process via differentiable soft constraints based on Lagrangian optimization, eliminating the need for sequential sub-modules.
    \item We construct CoTIR-Bench, a new large-scale benchmark, and demonstrate through extensive experiments that our method achieves SOTA performance across diverse degradation scenarios.
\end{itemize}


% \subsection{Subsection Heading Here}
% Subsection text here.

% % needed in second column of first page if using \IEEEpubid
% %\IEEEpubidadjcol

% \subsubsection{Subsubsection Heading Here}
% Subsubsection text here.

\section{Related Work}

\textbf{Diffusion and Flow Models.}
Diffusion models have achieved strong performance in image generation, with representative methods such as DDPM~\cite{ho2020denoising}, DDIM~\cite{song2020denoising}, LDM~\cite{rombach2022high}, and DiT~\cite{peebles2023scalable} steadily improving generation quality. However, their reliance on extensive iterative sampling incurs high computational costs. To address this, Flow Matching (FM)~\cite{lipman2022flow} reformulates the underlying stochastic differential equations into Ordinary Differential Equations (ODEs), enabling faster inference by learning straighter vector field trajectories.
FM’s core idea is further advanced by Rectified Flow~\cite{liu2022flow}, Shortcut Models~\cite{frans2024one}, and MeanFlows~\cite{geng2025mean}, which accelerate generation by optimizing more direct paths in the latent space. Building on FM, \texttt{FLUX.1}~\cite{batifol2025flux} and \texttt{FLUX.2}~\cite{flux-2-2025} achieve high-quality image generation and editing with minimal steps. Thus, we adopt \texttt{FLUX} model as our base model due to its strong editing capabilities and efficient inference.

\textbf{Generative Image Restoration.}
Conventional image restoration methods~\cite{zhu2023learning, kulkarni2022unified, potlapalli2024promptir, guo2024onerestore} learn deterministic mappings from degraded to clean images using small-scale paired datasets. These models often struggle with generalization due to limited training diversity and the inherently ill-posed nature of restoration tasks.
To overcome these limitations, recent works~\cite{chen2025adversarial, lin2024diffbir, wang2024exploiting, yang2024pixel, pu2025lumina, yu2024scaling, wu2024one} employ generative models pre-trained on large datasets. These models leverage learned priors to synthesize realistic details, improving robustness and visual quality. For instance, SUPIR~\cite{yu2024scaling} scales generative capacity to achieve photorealistic results, while OSEDiff~\cite{wu2024one} reduces the inference burden by introducing a one-step diffusion network. Our method builds upon this generative foundation, but emphasizes structured reasoning to better handle complex, composite degradations.

\textbf{Chain-of-Thought Reasoning.}
Chain-of-Thought (CoT) reasoning was originally introduced to enhance complex task solving in Large Language Models (LLMs)~\cite{wei2022chain}, where intermediate steps help decompose difficult problems. This paradigm has since been extended and refined~\cite{wang2023self, zhou2023least}, and more recently adapted to multimodal tasks, enabling models to reason visually through intermediate representations~\cite{su2025thinking, chern2025thinking, yang2025mmada}.
In image restoration, CoT-based approaches such as Q-Agent~\cite{zhou2025q}, Chain-of-Restoration~\cite{cao2024chain}, RestoreAgent~\cite{chen2024restoreagent}, and AgenticIR~\cite{zhu2024intelligent} apply this reasoning sequentially, invoking specialized modules to address different degradations step-by-step. While effective in decomposition, these methods are inefficient and fail to capture the intertwined nature of composite degradations. 
In contrast, our method internalizes the CoT process within a single-pass generative framework, enabling efficient and holistic restoration without relying on sequential decision-making.

\begin{figure*}[t]
    \centering
    \includegraphics[width=1\linewidth]{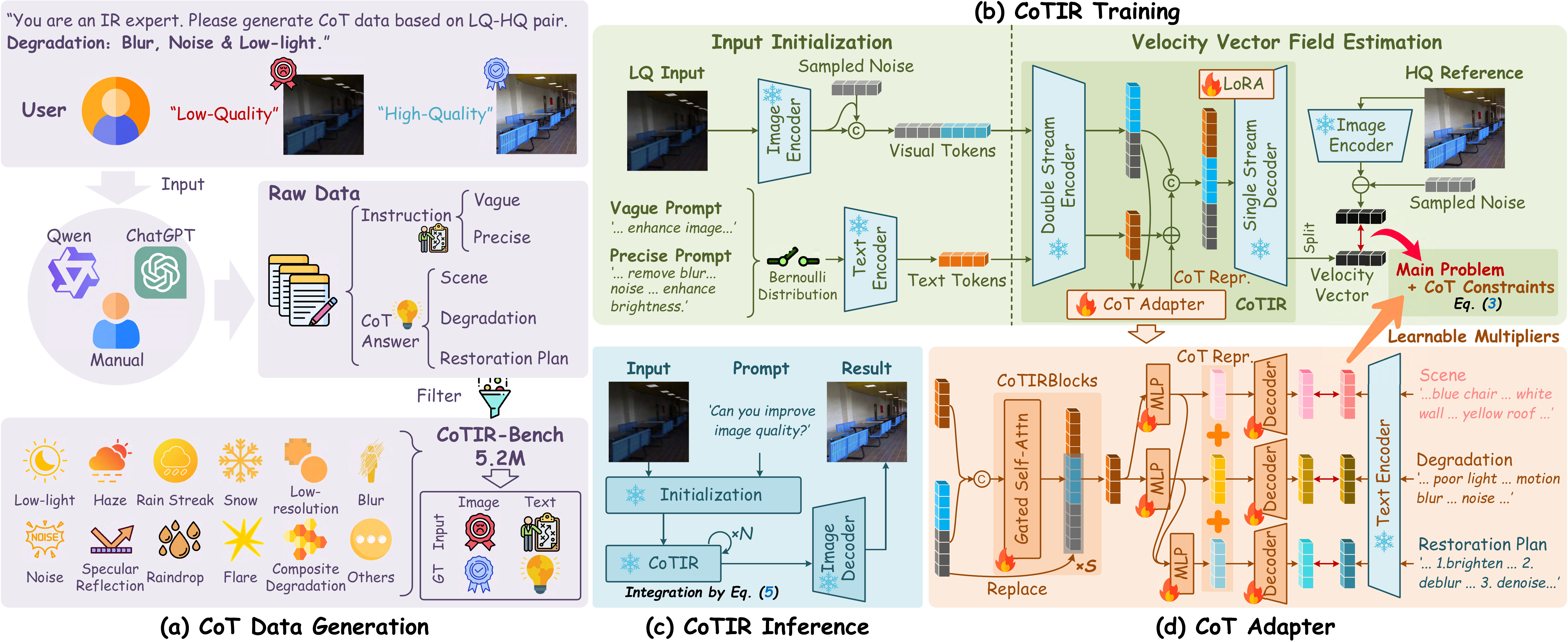} \\
    \vspace{-3mm}
    \caption{Architecture of our CoTIR. We employ advanced Vision-Language Models (VLMs) to generate high-quality CoT data for model training. Notably, a CoT Adapter module is specifically incorporated to internalize the VLM's step-by-step reasoning process into our fine-tuned restoration model.}
    \vspace{-6mm}
    \label{fig:cotir}
\end{figure*}

\section{Problem Formulation}

\subsection{Chain-of-Thought Reasoning in Image Restoration}
CoT reasoning has demonstrated critical efficacy in large language models by systematically guiding decision-making. We posit that an analogous structured rationale can also substantially benefit Image Restoration~(IR) confronted with intricate degradations.
Inspired by disentangling visual representations~\cite{saini2022disentangling}, we construct a pipeline tailored for IR, comprising the following steps:
\vspace{1mm}
\begin{center}
\begin{tcolorbox}[
  colback=gray!15,
  colframe=gray!15,
  arc=4pt,
  boxrule=0pt,
  left=2mm, right=2mm, top=1mm, bottom=1mm,
  width=\linewidth
]
$\bullet$ \textbf{Thinking--Feature Disentanglement:} learning intermediate representations of \textbf{\ding{182} inherent feature description} and \textbf{\ding{183} degradation pattern identification}.\\
$\bullet$ \textbf{Planning--Strategic Recovery Formulation:} analyzing the interplay between disentangled representations to define \textbf{\ding{184} restoration plan} for degradation removal.\\
$\bullet$ \textbf{Action--Guided Image Restoration:} leveraging \textbf{Restorer} to remove degradations and restore fined details.
\end{tcolorbox}
\end{center}

With the introduction of the CoT process, the image restoration task can be decomposed into a series of generation steps. 
Given a degraded input image $x$, the process first generates a set of textual intermediate results $\mathcal{C} = \{c_\text{s}, c_\text{d}, c_\text{p}\}$, where $c_\text{s}$ and $c_\text{d}$ represent sharp feature and degradation pattern descriptions, respectively, and $c_\text{p}$ is a restoration plan. The final restored image $\hat{y}$ is then generated based on $\mathcal{C}$ and $x$. 
The core task is to find an optimal mapping $f: x \to y$, where $y$ is the high-quality clean image. 
The CoT framework constrains this mapping to a structured reasoning path: $x \to \{c_\text{s}, c_\text{d}\} \to c_\text{p} \to y$. 
Thus, this problem can be formulated as a multi-constrained optimization problem, where the goal is to minimize the restoration error while ensuring that the intermediate outputs must align with their respective reasoning targets.

\subsection{Image Restoration Problem under CoT Constraints}
\label{sec:problem}
\textbf{Image Restoration Problem.}
We adopt \texttt{FLUX.1} Kontext as the base model and internalize CoT reasoning modes into its conditional generation process for robust restoration. 
The primary restoration task is formulated as learning a vector field $v_\theta$ that maps a random Gaussian noise $\varepsilon \sim \mathcal{N}(0, \mathbf{I})$ to the clean latent embedding $g$\footnote{$g$ is latent space representation of a clean image $y$ obtained by a VAE encoder.}, guided by the low-quality input image $x$, the textual prompt $c_0$, and the intermediate reasoning steps $\hat{\mathcal{C}}$. 
The training objective for the vector field is to minimize the flow matching loss:
\begin{equation}
\mathcal{L}_{\text{FM}}(\theta) = \mathbb{E}_{t, \varepsilon, g, c_0} \Big[ \| v(z_t, t, x, c_0, \hat{\mathcal{C}}; \theta) - (\varepsilon - g) \|_2^2 \Big],
\end{equation}
where $\theta$ denotes the model parameters, $t$ is the time step sampled from a logit-normal shift schedule, and $z_t$ is obtained via linear interpolation between $\varepsilon$ and $g$, i.e., $z_t = (1-t)g + t\varepsilon$.

\textbf{CoT Constraints.}
The CoT reasoning process introduces a set of intermediate textual constraints.
We require the model to produce a set of prediction results $\hat{\mathcal{C}} = \{\hat{c}_\text{s}, \hat{c}_\text{d}, \hat{c}_\text{p}\}$ that align with the ground-truth intermediate representations $\mathcal{C}$. 
Concretely, we impose tolerance constraints on their L2 deviations: $\lVert \hat{c}_{i}(z_t, t, x, c_0; \theta) - c_i \rVert_2^2 \le \delta_i$, $i=\text{s},\text{d},\text{p}$, where $\delta_i \ge 0$ denote per-constraint tolerances.

\textbf{Multi-Constrained Optimization.}
Consequently, the original restoration problem is transformed into the following multi-constraint optimization problem:
\begin{equation} \label{eq:constrained_opt}
\min_{\theta} \ \mathcal{L}_{\text{FM}}(\theta) \quad \text{s.t.} \, \lVert \hat{c}_{i}(z_t, t, x, c_0; \theta) - c_i \rVert_2^2 \le \delta_i, \ i = \text{s}, \text{d}, \text{p}.
\end{equation} 

By introducing the method of Lagrange multipliers, the constrained optimization problem can be reformulated as:
\begin{align} \label{eq:constrained_opt_lmm}
\mathcal{L}&(\theta, \lambda) = \mathbb{E}_{t, \varepsilon, g, c_0} \Big[  \| v(z_t, t, x, c_0, \hat{\mathcal{C}}; \theta) - (\varepsilon - g) \|_2^2 \nonumber \\
& \quad + \sum_{i \in \{\text{s, d, p}\}} \lambda_i \big( \lVert \hat{c}_{i}(z_t, t, x, c_0; \theta) - c_i \rVert_2^2 - \delta_i \big) \Big],
\end{align}
where $\lambda_i \ge 0$ are the Lagrange multipliers. This leads to the min-max formulation $\min_{\theta} \max_{\lambda \ge 0} \mathcal{L}(\theta, \lambda)$, which we optimize using a dual-optimizer primal--dual procedure. 
%More descriptions of this optimization are provided in Sec.~\ref{sec:training}, and its advantage over fixed coefficients is analyzed in Sec.~\ref{sec:theory_lar}.
%This min-max optimization is addressed using two separate optimizers. One updates the model parameters $\theta$, which comprise only the LoRA-fine-tuned parameters of the FLUX.1 decoder and the parameters of the CoT adapter, to minimize the objective. Concurrently, a second optimizer updates the Lagrange multipliers $\{\lambda_i\}$ to maximize the same objective while enforcing non-negativity. Further details on the optimization are provided in Sec. \ref{sec:training}.
\section{CoTIR}
\label{sec:cotir}

\subsection{CoT Data Generation}
As shown in Fig.~\ref{fig:cotir}a, we use Qwen2.5-VL \cite{bai2025qwen2} under a fixed instruction template that conditions on the degradation type text, the low-quality input $x$, and the high-quality ground-truth $y$ together with our designed instruction. The model then produces a high-quality, precise restoration instruction $\phi_\text{pre}$ and a CoT triple $\mathcal{C} = \{c_\text{s}, c_\text{d}, c_\text{p}\}$ comprising inherent feature description, degradation information, and restoration plan. For adaptive image restoration, we also prepare a set of vague prompts $\phi_\text{gen}$, created both manually and with GPT-5 Codex \cite{openai2024codex}. %Detailed templates and generation examples are provided in the Appendix \ref{sec:cotir-bench}. 
To ensure prompt quality, we filter the output, apply simple CLIP-based image-text consistency checks against $\phi_\text{pre}$, and retain high-quality prompts.

% Note that these CoT annotations are distilled from image cues by the vision-language model; the proposed CoT Adapter internalizes this thinking model and induces CoT reasoning implicitly during inference, without any external LLM or explicit CoT generation.

\subsection{CoTIR Architecture}
We build CoTIR upon \texttt{FLUX}-based editing models and further adapt them for image restoration. Fig.~\ref{fig:cotir}b-d illustrate the overall training and inference pipelines, together with the proposed CoT Adapter.

\textbf{Input Initialization.} 
The low-quality degraded image $x$ is encoded by the VAE encoder to obtain an image-conditioned embedding $h_x = \mathcal{E}_{\text{v}}(x)$.
To handle the text prompt selection, we introduce a staged Bernoulli schedule formalized as $\pi(k) = \min(0.2 + 0.1 \cdot \lfloor k/10^4 \rfloor, 0.6)$, where $b\sim\mathrm{Bernoulli}(\pi(k))$. 
The final text prompt is determined by $\phi = b\,\phi_{\text{gen}} + (1-b)\,\phi_{\text{pre}}$, and is then projected into the embedding space by the T5 text encoder $\mathcal{E}_{\text{t}}$ \cite{raffel2020exploring} to yield $c_0 = \mathcal{E}_{\text{t}}(\phi)$.
This progressive learning strategy implements a curriculum. Initially, the schedule assigns a higher probability to the precise prompt $\phi_{\text{pre}}$, allowing the model to learn fundamental restoration patterns. As training advances, the probability of selecting the general prompt $\phi_{\text{gen}}$ increases by 0.1 every 10k iterations. This gradually raises the training difficulty, compelling the model to infer degradation types rather than merely executing instructions, thereby building its reasoning capabilities.
Finally, we sample a time step $t$ from a logit-normal schedule and encode it via Fourier embedding $\gamma(t)$, then construct the latent input $z_t$ by interpolating between $\varepsilon$ and the clean latent embedding $g = \mathcal{E}_{\text{v}}(y)$ at time $t$.

\begin{figure}[t]
    \centering
    \includegraphics[width=1\linewidth]{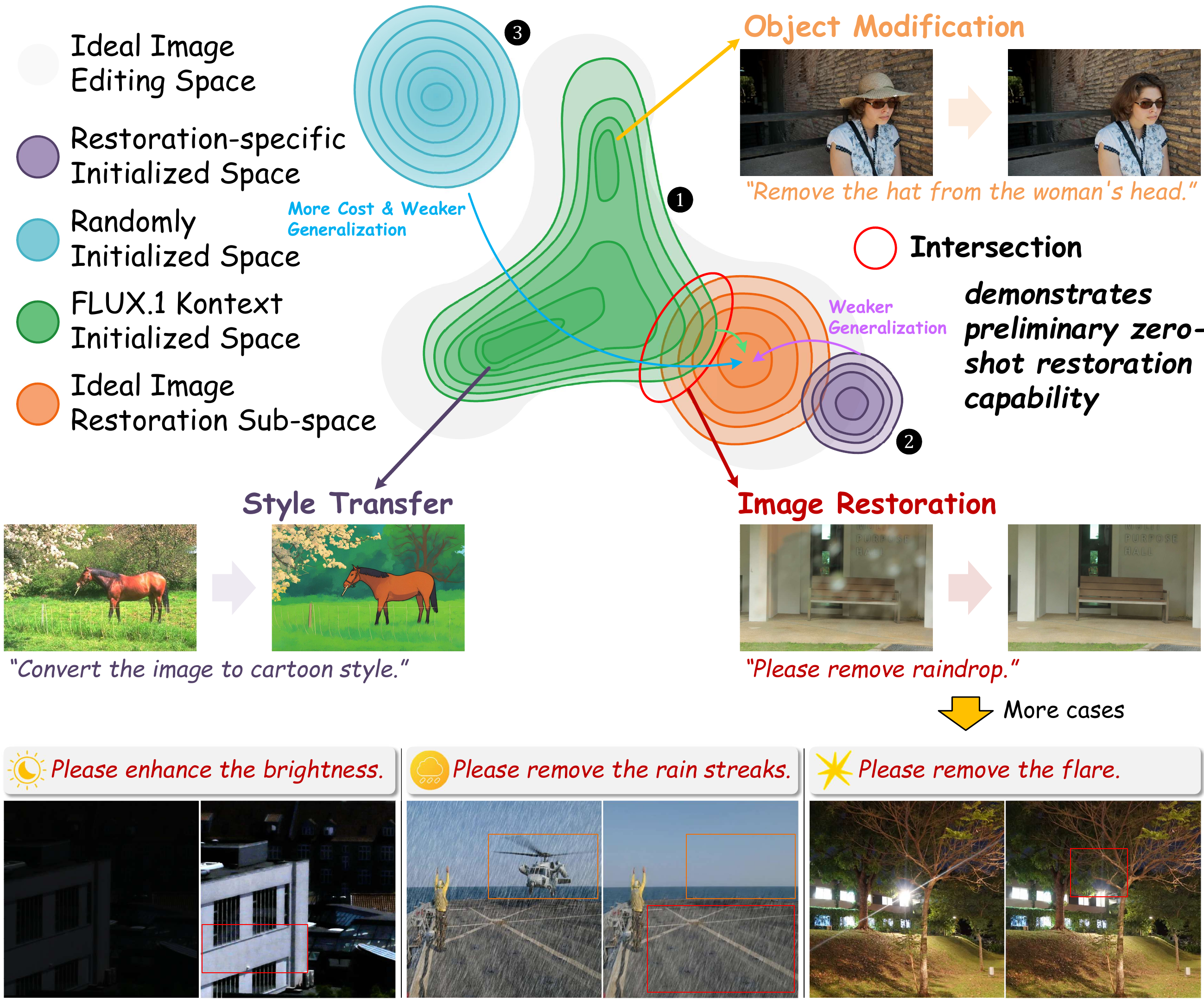} \\
    \vspace{-3mm}
    \caption{Comparison of three initialization strategies. The editing model already exhibits strong zero-shot restoration capability before fine-tuning, which motivates us to adopt editing models as initialization. Nevertheless, it still suffers from degradation remnants ({\color{red}red box}) and intrinsic feature/object variation ({\color{orange}orange box}), thus requiring further fine-tuning to better adapt the model to the target sub-operation space.}
    \label{fig:flux_potential}
    \vspace{-6mm}
\end{figure}

\textbf{Editing Prior.}
We conceptualize image restoration as a subspace of the broader image editing space. As illustrated in Fig.~\ref{fig:flux_potential}, a large-scale pre-trained editing model already exhibits strong zero-shot restoration ability on real-world scenes before task-specific fine-tuning. Fine-tuning then specializes this capability to restoration while retaining the broad priors from editing pre-training, yielding stronger generalization than both random initialization and restoration-specific pre-training, especially on realistic and unseen degradation combinations.

Mathematically, let $p_{\theta}(y\mid x,c)$ denote an editing model conditioned on an instruction $c$, where $x$ is the degraded input and $y$ is the restored image. A restoration task can be represented by a family of restoration instructions $c_{\mathcal{R}}$, which induces a restoration conditional model. For notational simplicity, we denote the resulting restoration conditional by $p_{\theta}(y\mid x)$ in the remainder of this subsection. The restoration objective is
\begin{equation}
\theta_{\mathcal{R}}^{\star}\in\arg\max_{\theta}\ \mathbb{E}_{(x,y)\sim D_{\mathcal{R}}}\big[\log p_{\theta}(y\mid x)\big],
\end{equation}
where $D_{\mathcal{R}}$ denotes the target restoration distribution. We compare three initializations. \ding{182} The editing initialization is obtained from large-scale editing pre-training,
\begin{equation}
\theta_{\text{edit}}\in\arg\max_{\theta}\ \mathbb{E}_{(x,c,y)\sim D_{\mathcal{J}}}\big[\log p_{\theta}(y\mid x,c)\big],
\end{equation}
where $D_{\mathcal{J}}$ covers broad image manipulation behaviors. \ding{183} The restoration-specific initialization is obtained from a narrower restoration benchmark,
\begin{equation}
\theta_{\text{ir}}\in\arg\max_{\theta}\ \mathbb{E}_{(x,y)\sim D_{\mathcal{B}}}\big[\log p_{\theta}(y\mid x)\big],
\end{equation}
where $D_{\mathcal{B}}$ contains limited restoration datasets. \ding{184} The third case is random initialization $\theta_{\text{rand}}$.

Initialization quality depends on the evaluation distribution. On the narrow benchmark $D_{\mathcal{B}}$, $\theta_{\text{ir}}$ can be highly competitive since it is optimized for those seen degradations. Universal restoration, however, targets a broader distribution with unseen combinations, spatially varying corruptions, and real-world cases whose degradation taxonomy is not fixed in advance. To analyze this regime, define the expected likelihood gap on $D_{\mathcal{R}}$ as
\begin{equation}
\Delta_{D_{\mathcal{R}}}(\theta_0)=\mathbb{E}_{(x,y)\sim D_{\mathcal{R}}}\big[\log p_{\theta_{\mathcal{R}}^{\star}}(y\mid x)-\log p_{\theta_0}(y\mid x)\big].
\end{equation}
Smaller $\Delta_{D_{\mathcal{R}}}(\theta_0)$ means that less adaptation is required to achieve a robust restorer. Under the universal restoration regime where $D_{\mathcal{R}}$ is substantially richer than $D_{\mathcal{B}}$, we expect the preferred initialization to satisfy
\begin{equation}
\Delta_{D_{\mathcal{R}}}(\theta_{\text{edit}})\le \Delta_{D_{\mathcal{R}}}(\theta_{\text{ir}})\le \Delta_{D_{\mathcal{R}}}(\theta_{\text{rand}}).
\end{equation}
This ordering reflects the expected advantage of editing initialization under the above distributional assumption. The restoration-specific initialization provides useful priors for seen degradations, but its likelihood model is still tied to a finite restoration taxonomy. In contrast, $\theta_{\text{edit}}$ is learned from a broader operation space $\mathcal{J}$ that contains restoration-like operations as a subset, and is therefore better aligned with target distributions involving unseen compositions, semantic preservation, and diverse scene structures.

\textbf{Model Finetuning.}
The \texttt{FLUX} encoder processes the joint visual input (concatenation of $h_x$ and $z_t$) and textual guidance $c_0$ through a double-stream architecture to produce latent embeddings $(\bar{V}, \bar{T})$. 
Subsequently, a CoT Adapter refines these embeddings to generate CoT features $\bar{T}_{\text{CoT}}$. 
Then, a single-stream decoder fuses $\bar{V}$ and $\bar{T}_{\text{CoT}}$ to predict the velocity field $v_{\theta}$. 
To reduce training costs while preserving the powerful generalization capability inherited from the editing prior, we only fine-tune the decoder via Low-Rank Adaptation (LoRA). 
For target weight matrix $W_m^0$ in the $m$-th linear layer in the decoder, we introduce trainable low-rank matrices $A_m$ and $B_m$ such that:
\begin{equation}
W_m = W_m^{0} + \frac{\alpha_m}{r} A_m B_m,
\end{equation}
where $\alpha_m$ is the scaling factor and $r$ is the rank of the low-rank matrices.
During training, only these low-rank matrices and the CoT Adapter parameters are optimized, while the original weights remain frozen.

\textbf{Model Inference.}
During inference, the restored image is generated by solving an Ordinary Differential Equation (ODE) to transform a noise latent $z_1 \sim \mathcal{N}(0, I)$ into a clean latent representation $z_0$. This is achieved by integrating the learned velocity field $v$ from $t=1$ to $0$:
\begin{equation}
z_0 = z_1 + \int_{1}^{0} v(z_t, t, x, c_0, \hat{\mathcal{C}}; \theta) \,dt.
\end{equation}
This integral is approximated numerically using a standard ODE solver. The resulting latent embedding $z_0$ is then mapped to the pixel space by the VAE decoder to produce the final output image $\hat{y}= \mathcal{D}_{\text{v}}(z_0)$.

\subsection{CoT Adapter} \label{sec:adapter}
The CoT Adapter $\mathcal{A}_{\text{CoT}}$ is conceptualized as a deep, multi-path module to emulate a structured reasoning process. It ingests the encoded text features $\bar{T}$ and visual features $\bar{V}$ to produce a CoT-enhanced text embedding, $\bar{T}_{\text{CoT}}$, which provides superior guidance for the restoration task.

\textbf{CoTIRBlock.} Inspired by GLIGEN \cite{li2023gligen}, the CoT adapter utilizes a stack of $S$ CoTIRBlocks to process the joint visual-textual input.
In each block, the visual embedding $\bar{V}$ incorporates text context $\bar{T}$ through a gated self-attention mechanism. 
Specifically, $\bar{T}$ and $\bar{V}$ are concatenated, modulated by the time embedding $\gamma(t)$, and processed by self-attention and MLP layers. 
Only text tokens are retained to update $\bar{T}$ via gated residual:
\begin{equation}
\begin{aligned}
    \Delta \bar{T} &= \mathrm{MLP}(\mathrm{Concat}(\bar{T}_{\text{Attn}}, \bar{T}_{\text{MLP}})), \\
    \bar{T}^{(s)} &= \bar{T}^{(s-1)} + \mathcal{G}(\gamma(t)) \odot \Delta \bar{T},
\end{aligned}
\end{equation}
where $\mathcal{G}(\gamma(t))$ is the gate function modulated by $\gamma(t)$. $\bar{T}_{\text{Attn}}$ and $\bar{T}_{\text{MLP}}$ denote the text-part features: $\bar{T}_{\text{Attn}}= \mathrm{Attn}(\dots)\big|_{1, \dots, N_t}$ and $\bar{T}_{\text{MLP}}=\mathrm{MLP}(\dots)\big|_{1, \dots, N_t}$.
This allows the text embedding to progressively query visual context for relevant information at each layer, enriching its semantic content without altering visual features.

\textbf{CoT Reasoning Internalization.} Finally, $\bar{T}^{(S)}$ is projected into three semantic sub-spaces: scene awareness $\bar{T}_{\text{s}}$, degradation identification $\bar{T}_{\text{d}}$, and restoration planning $\bar{T}_{\text{p}}$:

\begin{equation}
\begin{aligned}
    \bar{T}_{\text{s}} &= \mathrm{Proj}_\text{s}(\bar{T}^{(S)}), \quad \bar{T}_{\text{d}} = \mathrm{Proj}_\text{d}(\bar{T}^{(S)}), \\
    \bar{T}_{\text{p}} &= \mathrm{Proj}_\text{p}(\mathrm{Concat}(\bar{T}_{\text{s}}, \bar{T}_{\text{d}})).
\end{aligned}
\end{equation}
The recovery plan $\bar{T}_{\text{p}}$ aggregates information from the preceding two subspaces to simulate a chain-of-thought reasoning process. These features are then aggregated via zero-initialized gates to form the CoT-enhanced guidance:
\begin{equation}
\bar{T}_{\text{CoT}} = \bar{T} + \sigma_\text{s} \bar{T}_{\text{s}} + \sigma_\text{d} \bar{T}_{\text{d}} + \sigma_\text{p} \bar{T}_{\text{p}}.
\end{equation}
Moreover, the semantic vectors are projected to form intermediate representations $\hat{\mathcal{C}} = \{\hat{c}_{\text{s}}, \hat{c}_{\text{d}}, \hat{c}_{\text{p}}\}$ for supervision: $\hat{c}_{i} = \mathrm{Proj}_{i}(\bar{T}_{i}, \gamma(t)), \forall i \in \{\text{s, d, p}\}$.

\subsection{Model Training}\label{sec:training}
Similar to Cooper \cite{gallego2025cooper}, we train the model by solving the minimax problem in Eq. \eqref{eq:constrained_opt_lmm} using a dual-optimizer approach to alternately optimize parameters $\theta$ and Lagrangian multipliers $\lambda$.

\begin{table}[t]
\centering
\caption{Progressive training configurations of three backbones. Each entry is reported as patch/batch size. 0/20/40/60k represent the number of iterations.}
\vspace{-1mm}
\label{tab:train_schedule}
\setlength{\tabcolsep}{10pt}
\renewcommand{\arraystretch}{1.0}
\begin{tabular}{l|c|c|c|c}
\toprule
\textbf{Backbone} & \textbf{0k} & \textbf{20k} & \textbf{40k} & \textbf{60k} \\
\midrule
\texttt{FLUX.1-12B} & 128/48 & 256/40 & 384/24 & 512/16 \\
\texttt{FLUX.2-4B} & 128/96 & 256/80 & 384/48 & 512/32 \\
\texttt{FLUX.2-9B} & 128/48 & 256/32 & 384/16 & 512/8 \\
\bottomrule
\end{tabular}
\vspace{-4mm}
\end{table}

\begin{table*}[t]
\centering
\caption{Quantitative comparisons with fine-tuned methods on CoTIR-Bench. $^\dagger$ means the input of degradation prompts.}
\label{tab:restore_full}
\vspace{-3mm}
\setlength{\tabcolsep}{7.5pt}
\renewcommand{\arraystretch}{0.95}
\begin{tabular}{ll|ccccc|ccc}
\toprule
\multicolumn{1}{c}{\multirow{2}{*}{\textbf{Methods}}} & \multicolumn{1}{c|}{\multirow{2}{*}{\textbf{Venue}}} & \multicolumn{5}{c|}{\textbf{No Reference}} & \multicolumn{3}{c}{\textbf{Full Reference}} \\
\multicolumn{1}{c}{} & \multicolumn{1}{c|}{} & CLIP-IQA+$\uparrow$ & Q-Align$\uparrow$ & LIQE$\uparrow$ & MACLIP$\uparrow$ & CLIP-IQA$\uparrow$ & PSNR$\uparrow$ & SSIM$\uparrow$ & LPIPS$\downarrow$ \\
\bottomrule
\rowcolor{gray!20} \multicolumn{10}{l}{\textbf{CNN/Transformer-based Methods}} \\
\toprule
PromptIR~\cite{potlapalli2024promptir} & NeurIPS23 & 0.5343 & 3.2822 & 2.7011 & 0.5081 & 0.4314 & 22.78 & 0.8028 & 0.2325 \\
OneRestore$^\dagger$~\cite{guo2024onerestore} & ECCV24 & 0.5443 & 3.4235 & 2.9409 & 0.5081 & 0.4335 & 24.76 & \cellcolor{orange!20}0.8584 & 0.1676 \\
InstructIR~\cite{conde2024instructir} & ECCV24 & 0.5404 & 3.3228 & 2.7986 & 0.5066 & 0.4359 & 22.94 & 0.8070 & 0.2161 \\
InstructIR$^\dagger$~\cite{conde2024instructir} & ECCV24 & 0.5353 & 3.3667 & 2.7914 & 0.5029 & 0.4262 & \cellcolor{yellow!20}25.13 & \cellcolor{red!20}0.8603 & 0.1763 \\
\bottomrule
\rowcolor{gray!20} \multicolumn{10}{l}{\textbf{Diffusion/Flow-based Methods}} \\
\toprule
DA-CLIP~\cite{luo2023controlling} & ICLR24 & 0.5765 & 3.5149 & 3.1581 & 0.5119 & 0.4609 & 24.00 & 0.8171 & 0.1593 \\
HYPIR$^\dagger$~\cite{lin2025harnessing} & TOG25 & \cellcolor{yellow!20}0.6318 & 3.8116 & 3.5813 & 0.5434 & 0.5081 & 21.55 & 0.6930 & 0.1737 \\
LucidFlux~\cite{fei2025lucidflux} & ICLR26 & 0.5632 & 3.5314 & 3.3671 & 0.5135 & 0.4537 & 17.13 & 0.5364 & 0.3798 \\
\midrule
\cellcolor{blue!20}CoTIR (Flux.1-12B) & & 0.6115 & 3.7700 & 3.4767 & 0.5352 & 0.4961 & 23.73 & 0.7930 & 0.1368 \\
\cellcolor{blue!20}CoTIR (Flux.1-12B)$^\dagger$ & & 0.6137 & 3.7835 & 3.5136 & 0.5370 & 0.5011 & 24.36 & 0.8015 & 0.1243\\
\cellcolor{blue!20}CoTIR (Flux.2-4B) & & 0.6272 & \cellcolor{orange!20}3.8849 & \cellcolor{yellow!20}3.7061 & \cellcolor{red!20}0.5550 & \cellcolor{yellow!20}0.5291 & 24.52 & 0.8360 & 0.1206 \\
\cellcolor{blue!20}CoTIR (Flux.2-4B)$^\dagger$ & & 0.6262 & 3.8676 & 3.7016 & \cellcolor{orange!20}0.5547 & \cellcolor{red!20}0.5295 & \cellcolor{orange!20}25.19 & \cellcolor{yellow!20}0.8454 & \cellcolor{orange!20}0.1128 \\
\cellcolor{blue!20}CoTIR (Flux.2-9B) & & \cellcolor{orange!20}0.6319 & \cellcolor{red!20}3.8974 & \cellcolor{red!20}3.7474 & 0.5540 & 0.5260 & 24.61 & 0.8315 & \cellcolor{yellow!20}0.1143 \\
\cellcolor{blue!20}CoTIR (Flux.2-9B)$^\dagger$ & & \cellcolor{red!20}0.6335 & \cellcolor{yellow!20}3.8780 & \cellcolor{orange!20}3.7329 & \cellcolor{yellow!20}0.5546 & \cellcolor{orange!20}0.5291 & \cellcolor{red!20}25.34 & 0.8415 & \cellcolor{red!20}0.1023 \\
\bottomrule
\end{tabular}
\vspace{-4mm}
\end{table*}

\begin{figure}[t]
    \centering
    \includegraphics[width=1\linewidth]{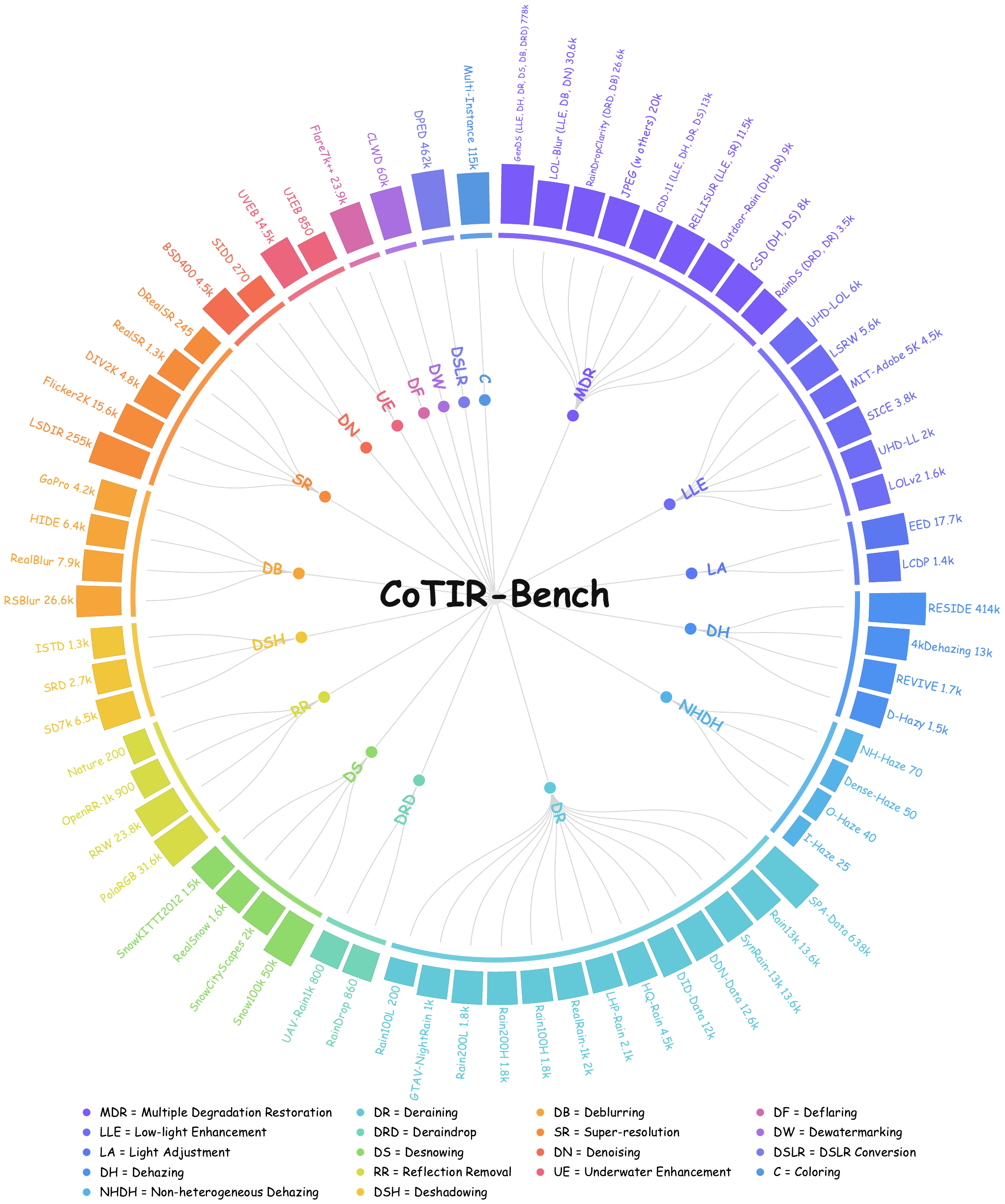}\\
    \vspace{-3mm}
    \caption{Composition of the CoTIR-Bench. The datasets and corresponding sample counts are shown in the enlarged view.}
    \label{fig:cotir-bench}
    \vspace{-5mm}
\end{figure}

\textbf{Parameter Optimization.} The primary optimizer minimizes the main loss with fixed $\lambda$ via stop-gradient $\mathrm{sg}(\cdot)$ \cite{van2017neural}:
\begin{equation}
\mathcal{L}_{\text{main}} = \| v - (\varepsilon - g) \|_2^2 + \sum_{i\in\{\text{s},\text{d},\text{p}\}} \mathrm{sg}(\lambda_i) ( \lVert \hat{c}_{i} - c_i \rVert_2^2 - \delta_i ).
\end{equation}

\textbf{Multiplier Optimization.} Concurrently, a second optimizer performs gradient ascent on $\lambda$ by minimizing:
\begin{equation}
\mathcal{L}_{\text{CoT}} = - \sum_{i\in\{\text{s},\text{d},\text{p}\}} \lambda_i \cdot \mathrm{sg}\big( \lVert \hat{c}_{i} - c_i \rVert_2^2 - \delta_i \big).
\end{equation}
% This strategy encourages the model to satisfy CoT constraints within margin $\delta_i$. The benefit of optimizing the multipliers as dual variables is that it explicitly targets the Lagrangian saddle point under the KKT conditions, whereas fixed coefficients correspond to a penalized objective that is usually more difficult to achieve the constrained optimum. 
% Our ablation study in Sec. \ref{sec:main_abl} further supports this points. 
% More details can be referred to Appendix \ref{sec:theory} and \ref{sec:method}.

\section{Experiments}
\subsection{Experiment Settings}
\textbf{Implementation Details.} 
Our framework is implemented in PyTorch and trained for 120k iterations on 8 NVIDIA A800 GPUs. We employ two AdamW optimizers with a shared configuration ($\beta_1=0.9, \beta_2=0.999, \text{weight decay}=0.01, \epsilon=10^{-8}$), using a learning rate of $5 \times 10^{-4}$ with cosine annealing for the main model and a fixed $1 \times 10^{-4}$ for the multipliers. Moreover, a progressive learning strategy is adopted to stabilize training. Since different backbones have different memories, we use different patch/batch configurations for \texttt{FLUX.1-12B}, \texttt{FLUX.2-4B}, and \texttt{FLUX.2-9B}, as summarized in Table~\ref{tab:train_schedule}.

\textbf{Datasets.}
Our training data consists of approximately 5.2 million high-quality data pairs, which is a large-scale aggregation of more than 60 publicly available datasets. This composite collection covers a wide spectrum of degradations, including Multiple Degradation Restoration (MDR), Low-Light Enhancement (LLE), Light Adjustment (LA), Dehazing (DH), Non-Heterogeneous Dehazing (NHDH), Deraining (DR), Deraindrop (DRD), Desnowing (DS), Reflection Removal (RR), Deshadowing (DSH), Deblurring (DB), Super-Resolution (SR), Denoising (DN), Underwater Enhancement (UE), and Deflaring (DF), Dewatermarking (DW), DSLR Conversion (DSLR), Coloring (C), and JPEG Repairing (JPEG). Fig. \ref{fig:cotir-bench} shows all used datasets and the number of samples in each dataset\footnote{The discrepancy between the original dataset counts and the final number of training pairs is due to data pre-processing and augmentation operations, such as cropping and resizing.}. To facilitate standardized evaluation, we have curated a unified test set by selecting 2,000 representative test samples from across these diverse datasets.

\begin{figure*}[t]
    \centering
    \includegraphics[width=1\linewidth]{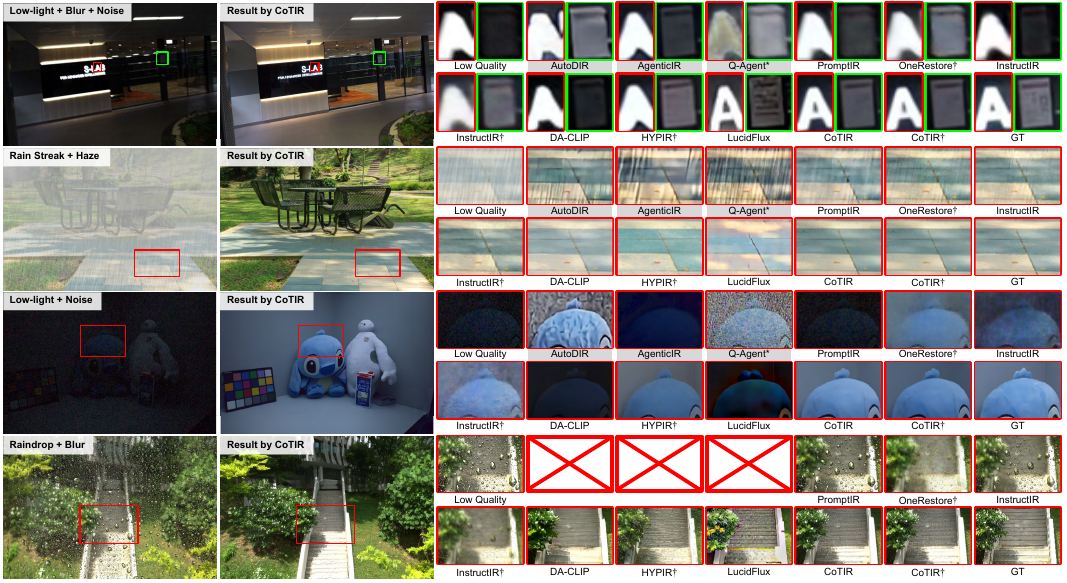}
    \vspace{-7mm}
    \caption{Composite degradation restoration comparisons on the CoTIR-Bench. \colorbox{gray!40}{Method} represents the results by corresponding official public checkpoints. $^\dagger$ means the input of degradation prompts.}
    \vspace{-5mm}
    \label{fig:results_composite}
\end{figure*}
\begin{table*}[t]
\centering
\caption{Quantitative comparisons with all-in-one methods on specific subtasks. $^\dagger$ means the input of degradation prompts.}
\vspace{-3mm}
\label{tab:restore_specific}
\setlength{\tabcolsep}{3.7pt}
\renewcommand{\arraystretch}{0.95}
\begin{tabular}{l|ll|ccccccccc}
\toprule
\multicolumn{1}{c|}{\textbf{Metrics}} & \textbf{Methods} & \textbf{Venue}  & \textbf{LLE} (127) & \textbf{DH} (117) & \textbf{DR} (409) & \textbf{DRD} (79) & \textbf{DS} (137) & \textbf{DB} (138) & \textbf{SR} (143) & \textbf{DN} (98) & \textbf{JPEG} (18) \\ 
\midrule
\multirow{7}{*}{CLIP-IQA+$\uparrow$} & InstructIR$^\dagger$~\cite{conde2024instructir} & ECCV24 & 0.5322 & 0.6293 & 0.5895 & - & - & 0.3967 & 0.5652 & 0.5748 & - \\
 & AutoDIR$^\dagger$~\cite{jiang2024autodir} & ECCV24 & \cellcolor{yellow!20}0.5800 & \cellcolor{yellow!20}0.6531 & 0.6217 & \cellcolor{yellow!20}0.6981 & - & 0.3938 & \cellcolor{orange!20}0.6929 & \cellcolor{yellow!20}0.6100 & - \\
 & DA-CLIP~\cite{luo2023controlling} & ICLR24 & 0.5668 & 0.6401 & \cellcolor{yellow!20}0.6242 & \cellcolor{yellow!20}0.6981 & \cellcolor{yellow!20}0.5854 & 0.3810 & - & 0.5535 & \cellcolor{yellow!20}0.5604 \\
 & DiffUIR$^\dagger$~\cite{zheng2024selective} & CVPR24 & 0.5366 & 0.6260 & 0.5976 & - & 0.5465 & 0.3401 & - & - & - \\
 & UniRestore~\cite{chen2025unirestore} & CVPR25 & - & 0.6025 & 0.6102 & - & 0.5389 & \cellcolor{yellow!20}0.4090 & - & 0.6037 & - \\
 & \cellcolor{blue!20}CoTIR & & \cellcolor{orange!20}0.6438 & \cellcolor{red!20}0.7000 & \cellcolor{red!20}0.6503 & \cellcolor{orange!20}0.7022 & \cellcolor{red!20}0.6070 & \cellcolor{red!20}0.5120 & \cellcolor{yellow!20}0.6704 & \cellcolor{red!20}0.6308 & \cellcolor{red!20}0.5955  \\
 & \cellcolor{blue!20}CoTIR$^\dagger$ & & \cellcolor{red!20}0.6480 & \cellcolor{orange!20}0.6969 & \cellcolor{orange!20}0.6481 & \cellcolor{red!20}0.7024 & \cellcolor{orange!20}0.6066 & \cellcolor{orange!20}0.5045 & \cellcolor{red!20}0.6999 & \cellcolor{orange!20}0.6223 & \cellcolor{orange!20}0.5941 \\
\midrule
\multirow{7}{*}{Q-Align$\uparrow$}& InstructIR$^\dagger$~\cite{conde2024instructir} & ECCV24 & 3.6442 & 3.5728 & 3.5248 & - & - & 2.9771 & 3.4505 & 3.2830 & - \\
 & AutoDIR$^\dagger$~\cite{jiang2024autodir} & ECCV24 & \cellcolor{yellow!20}3.7754 & \cellcolor{yellow!20}3.7220 & 3.6044 & \cellcolor{yellow!20}4.0952 & - & \cellcolor{yellow!20}3.1935 & \cellcolor{orange!20}4.2715 & \cellcolor{yellow!20}3.3213 & - \\
 & DA-CLIP~\cite{luo2023controlling} & ICLR24 & 3.6058 & 3.6672 & 3.5821 & 4.0946 & \cellcolor{yellow!20}3.6565 & 2.9483 & - & 3.2874 & \cellcolor{yellow!20}3.2534 \\
 & DiffUIR$^\dagger$~\cite{zheng2024selective} & CVPR24 & 3.5094 & 3.5227 & \cellcolor{yellow!20}3.6375 & - & 3.5859 & 2.7125 & - & - & - \\
 & UniRestore~\cite{chen2025unirestore} & CVPR25 & - & 3.6568 & 3.3484 & - & 3.3679 & 2.7214 & - & 3.2557 & - \\
 & \cellcolor{blue!20}CoTIR & & \cellcolor{orange!20}4.1978 & \cellcolor{red!20}4.1976 & \cellcolor{red!20}3.7861 & \cellcolor{red!20}4.3424 & \cellcolor{red!20}3.7548 & \cellcolor{red!20}3.6756 & \cellcolor{yellow!20}4.2701 & \cellcolor{red!20}3.7892 & \cellcolor{red!20}3.4134 \\
 & \cellcolor{blue!20}CoTIR$^\dagger$ & & \cellcolor{red!20}4.2059 & \cellcolor{orange!20}4.1338 & \cellcolor{orange!20}3.7580 & \cellcolor{orange!20}4.3369 & \cellcolor{orange!20}3.7447 & \cellcolor{orange!20}3.5182 & \cellcolor{red!20}4.4049 & \cellcolor{orange!20}3.7051 & \cellcolor{orange!20}3.2946 \\
\midrule
\multirow{7}{*}{PSNR$\uparrow$} & InstructIR$^\dagger$~\cite{conde2024instructir} & ECCV24 & 19.74 & 20.72 & \cellcolor{yellow!20}29.11 & - & - & \cellcolor{red!20}29.22 & \cellcolor{orange!20}27.19 & \cellcolor{red!20}34.22 & - \\
 & AutoDIR$^\dagger$~\cite{jiang2024autodir} & ECCV24 & 19.39 & 22.41 & \cellcolor{red!20}30.15 & \cellcolor{yellow!20}21.50 & - & 26.45 & 23.65 & \cellcolor{orange!20}33.91 & - \\
 & DA-CLIP~\cite{luo2023controlling} & ICLR24 & 17.63 & 22.52 & 27.38 & 21.24 & \cellcolor{orange!20}27.12 & 22.13 & - & 26.16 & \cellcolor{orange!20}29.35 \\
 & DiffUIR$^\dagger$~\cite{zheng2024selective} & CVPR24 & 18.35 & 22.13 & \cellcolor{orange!20}29.23 & - & \cellcolor{red!20}28.58 & \cellcolor{orange!20}28.18 & - & - & - \\
 & UniRestore~\cite{chen2025unirestore} & CVPR25 & \cellcolor{yellow!20}20.53 & \cellcolor{yellow!20}23.52 & - & 21.31 & 25.44 & - & \cellcolor{red!20}28.99 & - \\
 & \cellcolor{blue!20}CoTIR & & \cellcolor{orange!20}21.56 & \cellcolor{red!20}25.04 & 27.56 & \cellcolor{red!20}22.23 & 26.48 & 24.34 & \cellcolor{yellow!20}25.70 & 28.03 & \cellcolor{yellow!20}24.28 \\
 & \cellcolor{blue!20}CoTIR$^\dagger$ & & \cellcolor{red!20}22.76 & \cellcolor{orange!20}24.64 & 28.07 & \cellcolor{orange!20}22.08 & \cellcolor{yellow!20}26.68 & \cellcolor{yellow!20}26.87 & 25.42 & \cellcolor{yellow!20}29.40 & \cellcolor{red!20}29.42 \\
\midrule
\multirow{7}{*}{LPIPS$\downarrow$} & InstructIR$^\dagger$~\cite{conde2024instructir} & ECCV24 & 0.1681 & 0.1515 & 0.1450 & - & - & \cellcolor{orange!20}0.1555 & 0.2954 & 0.1143 & - \\
 & AutoDIR$^\dagger$~\cite{jiang2024autodir} & ECCV24 & \cellcolor{yellow!20}0.1599 & \cellcolor{yellow!20}0.1207 & \cellcolor{yellow!20}0.1148 & 0.2397 & - & 0.2131 & \cellcolor{yellow!20}0.1776 & \cellcolor{orange!20}0.0753 & - \\
 & DA-CLIP~\cite{luo2023controlling} & ICLR24 & 0.2014 & 0.1318 & 0.1521 & \cellcolor{yellow!20}0.2218 & \cellcolor{yellow!20}0.1174 & 0.2508 & - & 0.2145 & \cellcolor{orange!20}0.1190 \\
 & DiffUIR$^\dagger$~\cite{zheng2024selective} & CVPR24 & 0.2307 & 0.1486 & 0.1405 & - & 0.1225 & 0.2007 & - & - & - \\
 & UniRestore~\cite{chen2025unirestore} & CVPR25 & - & 0.1894 & 0.2729 & - & 0.2610 & 0.2220 & - & 0.1433 & - \\
 & \cellcolor{blue!20}CoTIR & & \cellcolor{orange!20}0.1045 & \cellcolor{red!20}0.0645 & \cellcolor{orange!20}0.0891 & \cellcolor{red!20}0.1265 & \cellcolor{orange!20}0.0773 & \cellcolor{yellow!20}0.1729 & \cellcolor{orange!20}0.1347 & \cellcolor{yellow!20}0.0894 & \cellcolor{yellow!20}0.1745 \\
 & \cellcolor{blue!20}CoTIR$^\dagger$ & & \cellcolor{red!20}0.0939 & \cellcolor{orange!20}0.0674 & \cellcolor{red!20}0.0848 & \cellcolor{orange!20}0.1269 & \cellcolor{red!20}0.0745 & \cellcolor{red!20}0.1180 & \cellcolor{red!20}0.1112 & \cellcolor{red!20}0.0748 & \cellcolor{red!20}0.0776 \\
\bottomrule
\end{tabular}
\vspace{-1mm}
\end{table*}

\begin{table*}[t]
\centering
\caption{Quantitative comparisons with multi-round restoration methods on CoTIR-Bench-7D (200). Q-Agent$^*$ only includes the computation time for the restorer and evaluator.}
\vspace{-3mm}
\label{tab:restore_7d}
\setlength{\tabcolsep}{7pt}
\renewcommand{\arraystretch}{0.95}
\begin{tabular}{ll|ccccc|ccc|l}
\toprule
\multicolumn{1}{c}{\multirow{2}{*}{\textbf{Methods}}} & \multicolumn{1}{c|}{\multirow{2}{*}{\textbf{Venue}}} & \multicolumn{5}{c|}{\textbf{No Reference}}                                                                                                      & \multicolumn{3}{c|}{\textbf{Full Reference}} & \textbf{Average}  \\
\multicolumn{1}{c}{}                         & \multicolumn{1}{c|}{}                       & CLIP-IQA+$\uparrow$ & Q-Align$\uparrow$ & LIQE$\uparrow$ & MACLIP$\uparrow$ & CLIP-IQA$\uparrow$ & PSNR$\uparrow$ & SSIM$\uparrow$ & LPIPS$\downarrow$ & \textbf{Runtime} \\
\midrule
AutoDIR~\cite{jiang2024autodir} & ECCV24 & \cellcolor{orange!20}0.5617 & \cellcolor{orange!20}3.3341 & \cellcolor{orange!20}3.1872 & \cellcolor{orange!20}0.5157 & \cellcolor{orange!20}0.4388 & \cellcolor{orange!20}19.96 & \cellcolor{yellow!20}0.6743 & \cellcolor{yellow!20}0.3020 &  327s \\
AgenticIR~\cite{zhu2024intelligent} & ICLR25 & \cellcolor{yellow!20}0.5039 & 3.0507 & 2.6099 & 0.4836 & 0.3979 & 16.98 & 0.6351 & 0.3366 & 55s \\
Q-Agent$^*$~\cite{zhou2025q} & Arxiv26 & 0.5003 & \cellcolor{yellow!20}3.2104 & \cellcolor{yellow!20}2.7110 & \cellcolor{yellow!20}0.4849 & \cellcolor{yellow!20}0.3801 & \cellcolor{yellow!20}19.83 & \cellcolor{orange!20}0.6956 & \cellcolor{orange!20}0.3010 & 39s$^*$\\
\cellcolor{blue!20}CoTIR & & \cellcolor{red!20}0.6559 & \cellcolor{red!20}4.1547 & \cellcolor{red!20}4.0984 & \cellcolor{red!20}0.5593 & \cellcolor{red!20}0.5429 & \cellcolor{red!20}22.82 & \cellcolor{red!20}0.7757 & \cellcolor{red!20}0.1350 & 4s \\
\bottomrule
\end{tabular}
\vspace{-3mm}
\end{table*}

\begin{figure}[t]
    \centering
    \includegraphics[width=1\linewidth]{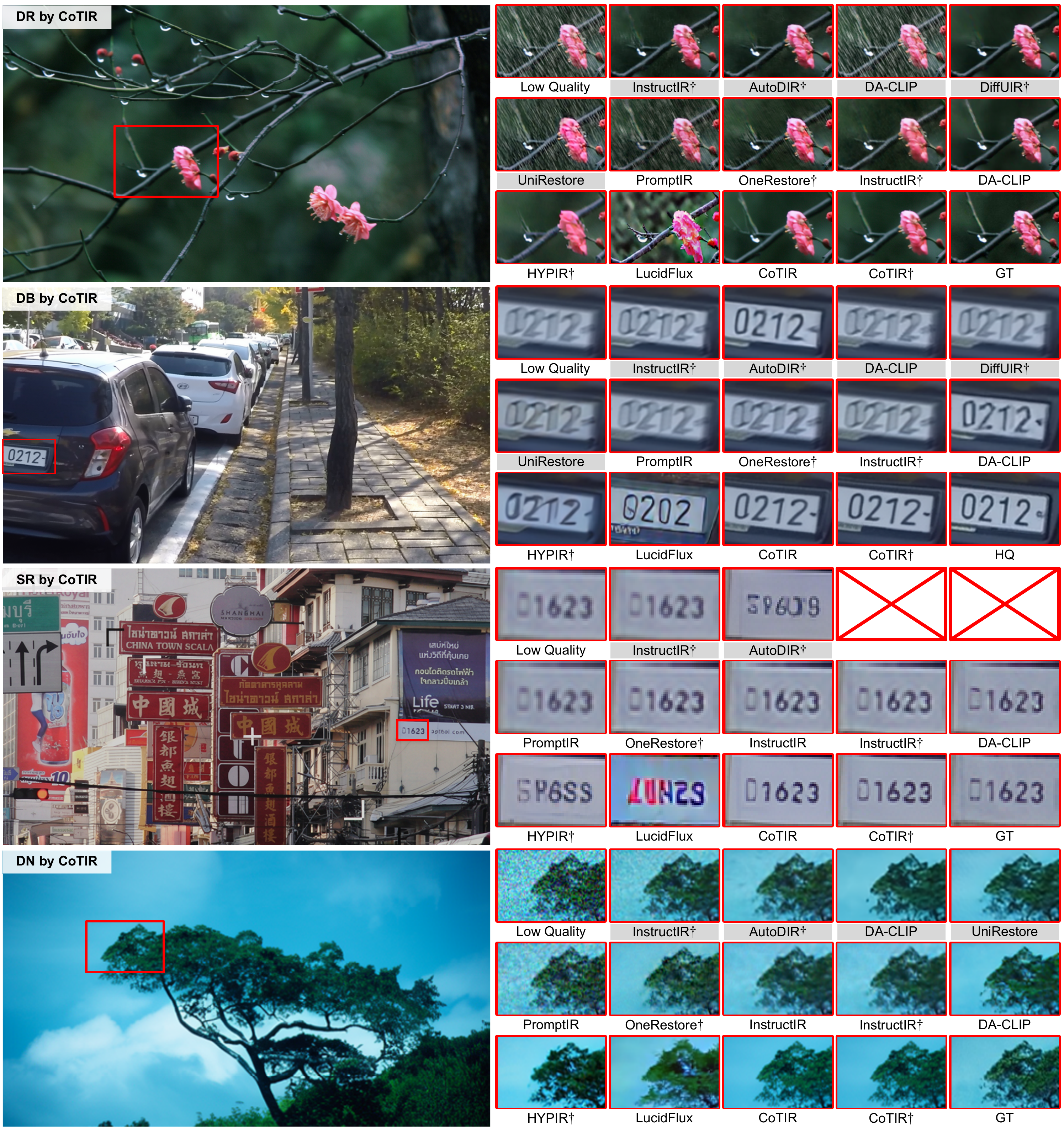}
    \vspace{-7mm}
    \caption{Single degradation restoration comparisons on the CoTIR-Bench. \colorbox{gray!40}{Method} represents the results by corresponding official public checkpoints. $^\dagger$ means the input of degradation prompts.}
    \vspace{-2mm}
    \label{fig:results_single}
\end{figure}

\begin{table}[t]
\caption{Quantitative comparisons on unseen AgenticIR Dataset~\cite{jiang2024autodir}. $^\dagger$ means the input of degradation prompts.}
\vspace{-3mm}
\label{tab:restore_agenticir}
\setlength{\tabcolsep}{0.95pt}
\resizebox{\linewidth}{!}{
\begin{tabular}{ll|ccc|ccc}
\toprule
\multicolumn{1}{c}{\multirow{2}{*}{\textbf{Methods}}} & \multicolumn{1}{c|}{\multirow{2}{*}{\textbf{Venue}}} & \multicolumn{3}{c|}{\textbf{No Reference}} & \multicolumn{3}{c}{\textbf{Full Reference}} \\
\multicolumn{1}{c}{} & \multicolumn{1}{c|}{} & MANIQA$\uparrow$ & CLIP-IQA$\uparrow$ & MUSIQ$\uparrow$ & PSNR$\uparrow$ & SSIM$\uparrow$ & LPIPS$\downarrow$ \\
\midrule
AirNet~\cite{li2022all} & CVPR22 & 0.2463 & 0.3799 & 40.54 & 18.85 & 0.5928         & 0.4529 \\
PromptIR$^\dagger$~\cite{potlapalli2024promptir} & NeurIPS23 & 0.2503 & 0.3837 & 40.56 & 19.74 & 0.6003 & 0.4369 \\
MiOIR$^\dagger$~\cite{kong2024towards} & Arxiv24 & 0.2299 & 0.3754 & 46.20 & 19.32 & 0.6193 & 0.4045 \\
InstructIR$^\dagger$~\cite{conde2024instructir}  & ECCV24 & 0.2503 & 0.3326 & 43.85 & 17.86 & 0.5714 & 0.4650 \\
AutoDIR~\cite{jiang2024autodir} & ECCV24 & 0.2383 & 0.3598 & 45.20 & 19.43 & 0.6151 & 0.4141 \\
DA-CLIP~\cite{luo2023controlling} & ICLR24 & 0.2283 & 0.3959 & 40.44 & 19.01 & 0.5810 & 0.4602 \\
AgenticIR~\cite{zhu2024intelligent} & ICLR25 & \cellcolor{yellow!20}0.3004 & \cellcolor{yellow!20}0.4376 & \cellcolor{orange!20}54.79 & \cellcolor{orange!20}20.29 & \cellcolor{red!20}0.6498 & \cellcolor{yellow!20}0.3501 \\
\cellcolor{blue!20}CoTIR & & \cellcolor{orange!20}0.3282 & \cellcolor{orange!20}0.4835 & \cellcolor{yellow!20}53.93 & \cellcolor{yellow!20}19.91 & \cellcolor{yellow!20}0.6383 & \cellcolor{orange!20}0.3098 \\
\cellcolor{blue!20}CoTIR$^\dagger$ & & \cellcolor{red!20}0.3348 & \cellcolor{red!20}0.4978 & \cellcolor{red!20}55.23 & \cellcolor{red!20}20.38 & \cellcolor{orange!20}0.6496 & \cellcolor{red!20}0.2932 \\
\bottomrule
\end{tabular}}
\vspace{-6mm}
\end{table}

\textbf{Comparison Methods and Evaluation Metrics.}
On our CoTIR-Bench, we fine-tune PromptIR~\cite{potlapalli2024promptir}, OneRestore~\cite{guo2024onerestore}, InstructIR~\cite{conde2024instructir}, DA-CLIP~\cite{luo2023controlling}, HYPIR~\cite{lin2025harnessing}, and LucidFlux~\cite{fei2025lucidflux} for comprehensive comparisons. We also compare with public checkpoints of InstructIR~\cite{conde2024instructir}, AutoDIR~\cite{jiang2024autodir}, DA-CLIP~\cite{luo2023controlling}, DiffUIR~\cite{zheng2024selective}, UniRestore~\cite{chen2025unirestore}, AgenticIR~\cite{zhu2024intelligent}, and Q-Agent$^*$\footnote{Since Q-Agent is not publicly available, we used its restorer and evaluator, combined with real degradation labels as CoT results, to simulate its ideal upper limit performance.}~\cite{zhou2025q} on the degradations considered by these methods. We use PSNR, SSIM, and LPIPS as full-reference metrics, and CLIP-IQA+, Q-Align, LIQE, MACLIP, and CLIP-IQA as no-reference metrics, where the higher, the better, except for LPIPS.

\subsection{Comparisons on CoTIR-Bench}

\textbf{Quantitative Comparisons.}
As shown in Table~\ref{tab:restore_full}, CoTIR achieves the best or near the best no-reference scores and the lowest LPIPS among the compared fine-tuned methods, while remaining competitive in PSNR and SSIM. This indicates that CoTIR is particularly effective for perceptual restoration quality under complex degradations, even though pixel-aligned metrics may favor some CNN/Transformer baselines in certain settings. Table~\ref{tab:restore_specific} shows a similar trend: while PSNR is slightly lower than the best baseline on several subtasks, CoTIR obtains clearly better perceptual results under no-reference metrics and LPIPS on most degradations. 
This discrepancy mainly arises because many target images still contain mild residual degradations, whereas our method tends to remove such degradations more thoroughly; consequently, pixel-wise metrics such as PSNR and SSIM, which rely on strict pixel alignment, may become less favorable, while perceptual metrics better capture the actual visual improvement produced by CoTIR. 

In addition, we construct a CoTIR-Bench-7D subset containing 200 images that cover seven degradations, including LLE, DH, DR, DB, SR, DN, and JPEG, as well as their combinations, for comparison with CoT-based multi-round restoration methods. These methods are typically system-level pipelines that select and compose existing restoration tools based on evaluation metrics, rather than learned restoration models with a unified optimization objective. As shown in Table~\ref{tab:restore_7d}, the advantage of CoTIR mainly comes from unified modeling of coupled degradations rather than better selection of restoration strategies, since sequential tool invocation inherently overlooks degradation coupling and often causes error accumulation across stages; consequently, CoTIR achieves both better overall performance and substantially higher efficiency.

\textbf{Qualitative Comparisons.}
Fig.~\ref{fig:results_composite} presents visual comparisons on composite degradation samples from CoTIR-Bench, where we use CoTIR (Flux.2-9B) as the default configuration for visualization. Multi-round restoration methods, such as AutoDIR, AgenticIR, and Q-Agent, struggle to handle real composite degradations effectively, since they decompose multiple degradations into sequential sub-problems and thereby overlook the strong coupling among them, often leaving noticeable residual artifacts. Finetuned restoration baselines also show clear limitations, with the large variety and combinations of degradations substantially increasing the difficulty of joint training. In particular, diffusion- or flow-based methods, such as DA-CLIP, HYPIR, and LucidFlux, usually exhibit better visual quality than CNN/Transformer-based methods despite their relatively weaker full-reference scores, yet some of them still behave unstably and may hallucinate non-existent details under severe degradations to make the outputs appear plausible. In contrast, the proposed method shows stronger scene consistency and more robust restoration behavior. Moreover, with CoT explicitly incorporated during training, CoTIR can stably perceive and restore the corresponding degradations even without being given degradation prompts. Fig.~\ref{fig:results_single} further focuses on several degradations where the full-reference PSNR in Table~\ref{tab:restore_specific} is not dominant, including DR, DB, SR, and DN. The zoomed-in results clearly show that CoTIR removes these degradations more robustly. For example, in the DN case, although InstructIR and AutoDIR achieve notably higher PSNR than our method, their outputs still fail to suppress noise effectively, whereas the result of CoTIR is visually closer to the GT. This observation is also consistent with the stronger performance of CoTIR on the other three perceptual metrics in Table~\ref{tab:restore_specific}.

\begin{table}[t]
\centering
\caption{Non-reference quantitative comparisons on the real-world composite degradation dataset.}
\vspace{-2mm}
\label{tab:real}
\setlength{\tabcolsep}{1pt}
\resizebox{\linewidth}{!}{
\begin{tabular}{ll|cccccc}
\toprule
\textbf{Methods} & \textbf{Venue} & \textbf{CLIP-IQA+$\uparrow$} & \textbf{Q-Align$\uparrow$} & \textbf{LIQE$\uparrow$} & \textbf{MACLIP$\uparrow$} & \textbf{CLIP-IQA$\uparrow$} & \textbf{US$\downarrow$} \\
\midrule
AutoDIR~\cite{jiang2024autodir} & ECCV24 & 0.5401 & 3.1978 & 2.5740 & \cellcolor{yellow!20}0.5093 & \cellcolor{yellow!20}0.4628 & 19.40 \\ 
AgenticIR~\cite{zhu2024intelligent} & ICLR25 & 0.5209 & 3.2034 & 2.5557 & 0.4955 & 0.4338 &27.50 \\
Q-Agent$^*$~\cite{zhou2025q} & Arxiv26 & 0.5223 & 3.1276 & 2.4639 & 0.4931 & 0.4415 & 17.20 \\
\midrule
PromptIR~\cite{potlapalli2024promptir} & NeurIPS23 & 0.5194 & 3.1447 & 2.4136 & 0.4947 & 0.4282 & 19.00 \\
OneRestore~\cite{guo2024onerestore} & ECCV24 & \cellcolor{yellow!20}0.5304 & \cellcolor{yellow!20}3.2733 & \cellcolor{yellow!20}2.6303 & 0.4822 & 0.4067 & \cellcolor{yellow!20}33.20 \\
InstructIR~\cite{conde2024instructir} & ECCV24 & 0.5063 & 3.1365 & 2.3328 & 0.4753 & 0.3948 & 25.80 \\
DA-CLIP~\cite{luo2023controlling} & ICLR24 & 0.5330 & 3.1735 & 2.5722 & 0.4861 & 0.4366 & 26.30 \\
HYPIR~\cite{lin2025harnessing} & TOG25 & \cellcolor{orange!20}0.5946 & \cellcolor{orange!20}3.5901 & \cellcolor{orange!20}3.1020 & \cellcolor{orange!20}0.5120 & \cellcolor{orange!20}0.4699 & \cellcolor{orange!20}55.10 \\
LucidFlux~\cite{fei2025lucidflux} & ICLR26 & 0.5310 & 3.1876 & 2.7699 & 0.4947 & 0.4224 & 
28.50 \\
\cellcolor{blue!20}CoTIR & & \cellcolor{red!20}0.6085 & \cellcolor{red!20}3.5943 & \cellcolor{red!20}3.2435 & \cellcolor{red!20}0.5381 & \cellcolor{red!20}0.5127 & \cellcolor{red!20}85.80 \\
\bottomrule
\end{tabular}}
\vspace{-3mm}
\end{table}

\begin{figure}[t]
    \centering
    \includegraphics[width=1\linewidth]{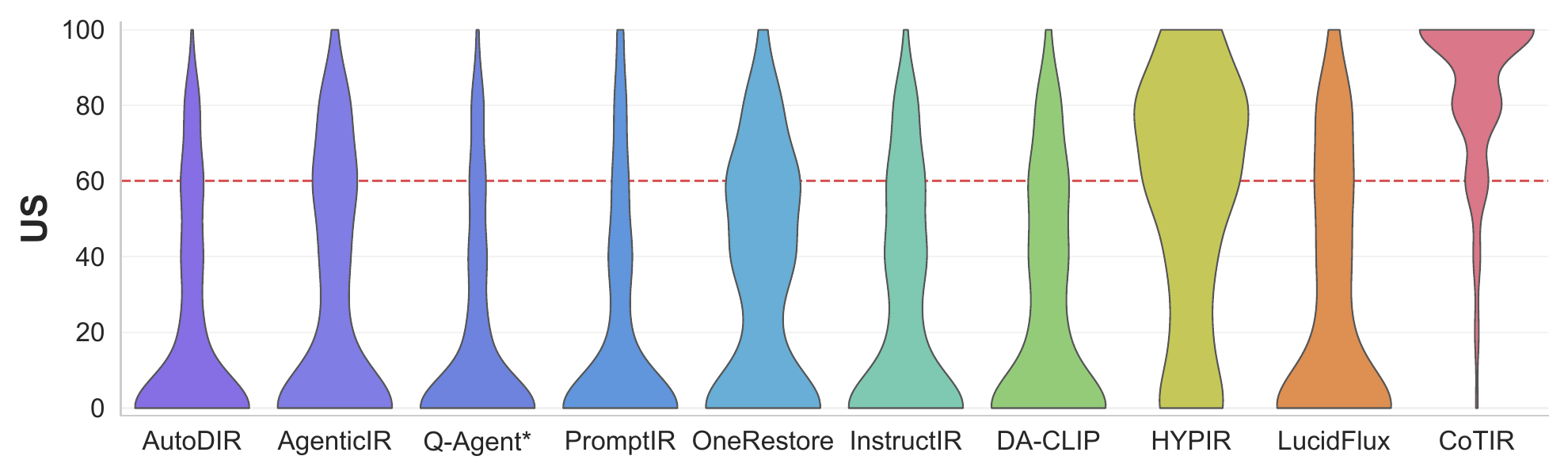}\\
    \vspace{-4mm}
    \caption{Comparison of user study for various restoration models. Our method achieves the highest perceptual scores.}
    \label{fig:vilon}
    \vspace{-1mm}
\end{figure}

\begin{figure*}[ht]
    \centering
    \includegraphics[width=1\linewidth]{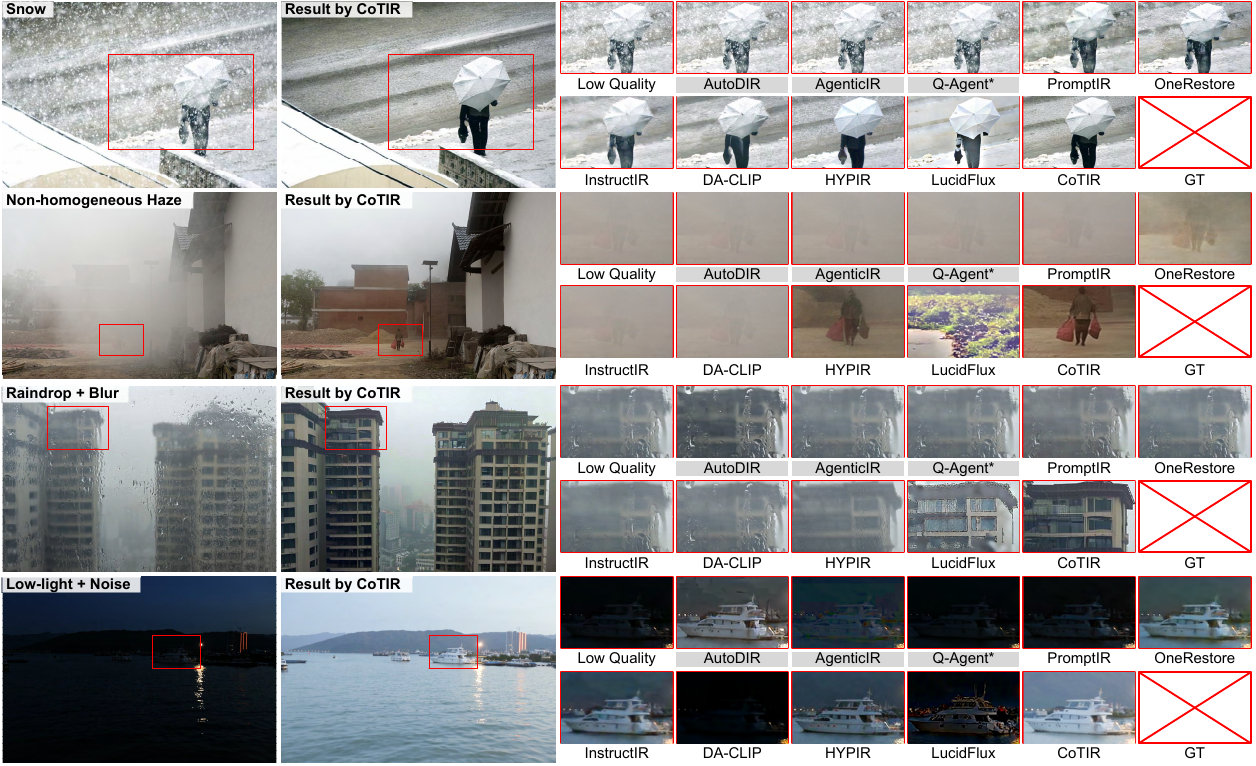}
    \vspace{-7mm}
    \caption{Visual comparisons of our method with others on the real scenes containing seen degradation. \colorbox{gray!40}{Method} represents the results by corresponding official public checkpoints. $^\dagger$ means the input of degradation prompts.}
    \vspace{-4mm}
    \label{fig:results_real_seen}
\end{figure*}

\begin{figure*}[htpb]
    \centering
    \includegraphics[width=1\linewidth]{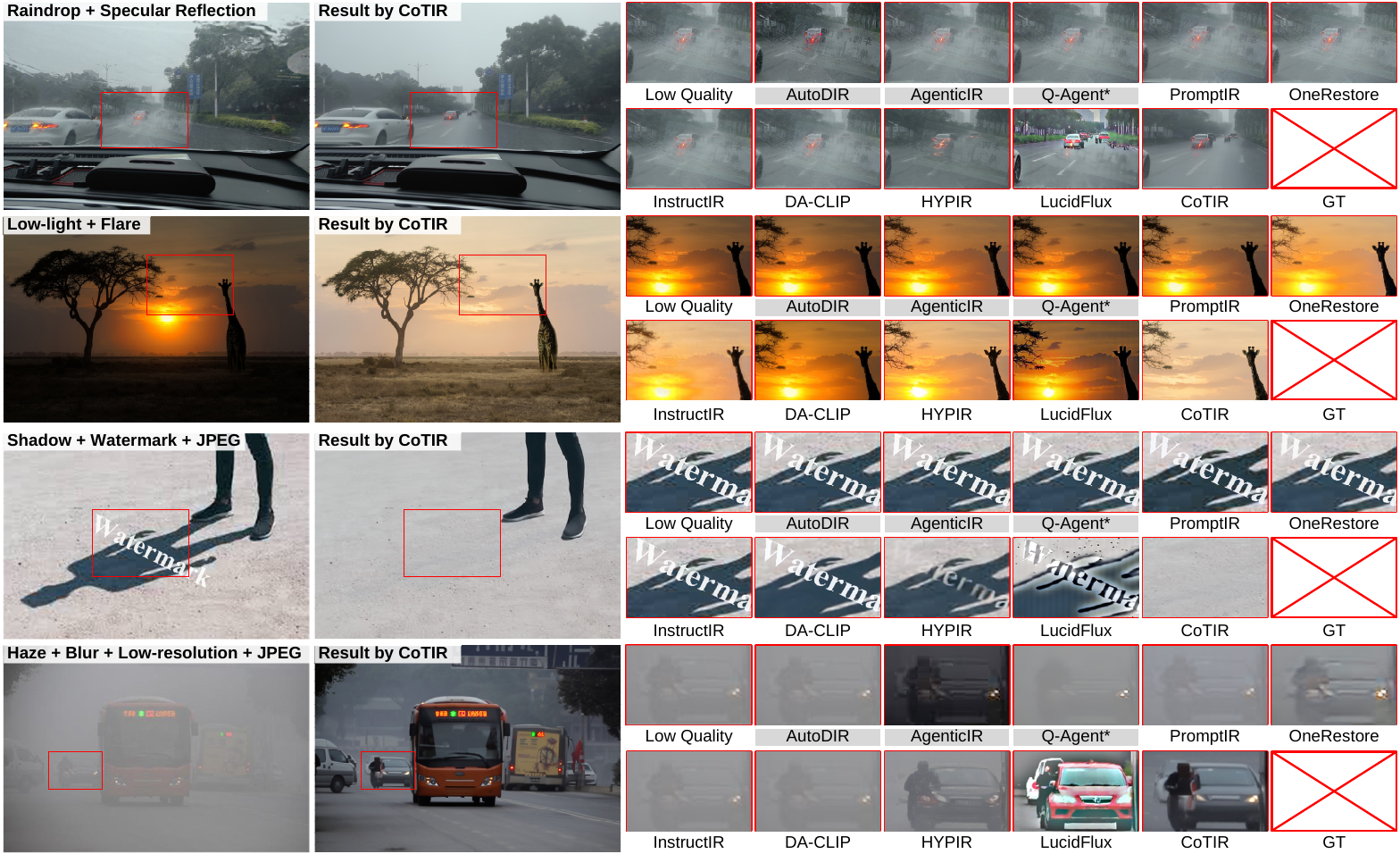}
    \vspace{-7mm}
    \caption{Visual comparisons of our method with others on the real scenes containing unseen degradation combination. \colorbox{gray!40}{Method} represents the results by corresponding official public checkpoints. $^\dagger$ means the input of degradation prompts.}
    \vspace{-5mm}
    \label{fig:results_real_unseen}
\end{figure*}

\begin{table}[t]
\centering
\caption{Ablation study on the CoTIR(Flux.1-12B) module configuration. `Vague' denotes generic user prompts, while `Precise' indicates accurate degradation recovery commands.}
\vspace{-2mm}
\label{tab:ab_cot}
\setlength{\tabcolsep}{1pt}
\resizebox{\linewidth}{!}{
\begin{tabular}{c|cc|cc|cc|cc|cc}
\toprule
\multirow{2}{*}{\textbf{No.}} & \multicolumn{2}{c|}{\textbf{Attention}} & \multicolumn{2}{c|}{\textbf{CoT Constraint}} & \multicolumn{2}{c|}{\textbf{CoT Coefficients}} & \multicolumn{2}{c|}{\textbf{Vague}}  & \multicolumn{2}{c}{\textbf{Precise}} \\
& CA & GA & Union & Split & Fixed & Learnable & PSNR$\uparrow$ & LPIPS$\downarrow$ & PSNR$\uparrow$ & LPIPS$\downarrow$ \\
\midrule 
(a) & & & & & & & 20.35 & 0.2013 & 21.26 & 0.2021 \\
(b) & \checkmark & & & & & & 21.72 & \cellcolor{yellow!20}0.1610 & 21.97 & 0.1561 \\
(c) & & \checkmark & & & & & \cellcolor{yellow!20}22.01 & 0.1679 & 22.57 & 0.1543 \\
 \midrule
(d) & & \checkmark& \checkmark & & & \checkmark & 21.88 & 0.1643 & \cellcolor{yellow!20}22.77 & \cellcolor{yellow!20}0.1484 \\
(e) & & \checkmark & &  \checkmark & \checkmark & &  \cellcolor{orange!20}22.75 & \cellcolor{orange!20}0.1469 & \cellcolor{orange!20}22.94 & \cellcolor{orange!20}0.1441 \\
(f) & & \checkmark & & \checkmark & & \checkmark & \cellcolor{red!20}23.73 & \cellcolor{red!20}0.1368 & \cellcolor{red!20}24.36 & \cellcolor{red!20}0.1243 \\
\bottomrule
\end{tabular}
}
\vspace{-4mm}
\end{table}

% \begin{table}[h]
% \centering
% \caption{Ablation study on the progressive training. `Vague' denotes generic user prompts, while `Precise' indicates accurate degradation recovery commands.}
% \label{tab:abl_pro}
% \setlength{\tabcolsep}{1pt}
% \resizebox{\linewidth}{!}{
% \begin{tabular}{l|cc|cc}
% \toprule
% \multirow{2}{*}{\textbf{Methods}} & \multicolumn{2}{c|}{\textbf{Vague}} & \multicolumn{2}{c}{\textbf{Precise}} \\
%  & PSNR$\uparrow$ & LPIPS$\downarrow$ & PSNR$\uparrow$ & LPIPS$\downarrow$ \\
% \midrule
% Baseline (w/o Progressive Training) & \cellcolor{yellow!20}21.31 & \cellcolor{yellow!20}0.2140 & \cellcolor{yellow!20}21.72 & \cellcolor{yellow!20}0.1920 \\
% \hspace{1em} + Patch/Batch Scheduling & \cellcolor{orange!20}23.12 & \cellcolor{orange!20}0.1481 & \cellcolor{orange!20}23.38 & \cellcolor{orange!20}0.1362 \\
% \hspace{2em} + Prompt Scheduling & \cellcolor{red!20}23.73 & \cellcolor{red!20}0.1368 & \cellcolor{red!20}24.36 & \cellcolor{red!20}0.1243 \\
% \bottomrule
% \end{tabular}
% }
% \end{table}

\subsection{Comparisons on Unseen AgenticIR Dataset.}

Table~\ref{tab:restore_agenticir} reports results on the unseen AgenticIR dataset, which contains degradation combinations not observed during training. Although our method is not specifically adapted to this dataset, CoTIR still outperforms the dedicated AgenticIR baseline on most no-reference and full-reference metrics. These results further verify the advantage of the proposed method in unified modeling of composite degradations. In contrast to multi-round restoration strategies, our method handles multiple degradations within a single model, leading to both better effectiveness and higher efficiency.

\subsection{Comparisons on Real-World Scenes.}

We further conduct real-world composite image restoration experiments on 350 samples collected from multiple real degradation datasets, including Yang's rain streak dataset \cite{yang2017deep}, NH-Haze21/23 non-homogeneous haze dataset \cite{ancuti2020nh}, RESIDE-RTTS haze dataset \cite{li2018benchmarking}, Snow100k snow dataset \cite{liu2018desnownet}, USR shadow dataset \cite{wang2018stacked}, BDN specular reflection dataset \cite{eccv18refrmv}, RUIE underwater dataset \cite{liu2020real}, et al. As shown in Table~\ref{tab:real}, we mainly report no-reference metrics and a user study with 30 volunteers. Each participant is asked to evaluate 20 samples randomly selected from the 350 real images and score the restored result of each method, resulting in 200 scores per participant. Before the study, the participants are explicitly instructed to assign a score of 60 or lower when the restoration is ineffective or visually unsatisfactory, and the available scores are restricted to six levels: 0, 20, 40, 60, 80, and 100. The results clearly show that the proposed method achieves the best overall performance under both no-reference metrics and human evaluation. In addition, the violin chart in Fig. \ref{fig:vilon} shows the score distribution of various methods, where the proportion of CoTIR samples scoring above 60 is significantly higher than that of other methods, demonstrating the effectiveness and stability. Fig.~\ref{fig:results_real_seen} shows qualitative comparisons on challenging real samples with seen degradation combinations, while Fig.~\ref{fig:results_real_unseen} further presents restoration results on real samples with unseen degradation combinations during training. In both settings, CoTIR effectively removes diverse degradation interference and produces clearer and more visually pleasing restoration results, demonstrating strong robustness and generalization on real composite degradations.

\subsection{Ablation Study}
\label{sec:main_abl}

We conduct ablation studies on the CoTIR-Bench test dataset based on Flux.1-12B to evaluate the proposed components, using PSNR and LPIPS as the full-reference metrics.

\textbf{Selection of Different Attention Mechanisms.}
We first investigate the impact of different adapter architectures. As presented in Table~\ref{tab:ab_cot} (a-c), the incorporation of an adapter structure significantly enhances performance compared to the baseline (a), validating the necessity of the adapter for task adaptation.
Regarding the attention mechanism, Cross-Attention (b) typically utilizes text features as the query and visual features as keys and values to enhance the text representation.
In contrast, Gated-Attention (c) treats image and text tokens as a unified whole for joint self-attention. This strategy facilitates deep multimodal fusion, allowing visual features to effectively guide the enhancement of textual representations.
Comparison between (b) and (c) reveals that the Gated-Attention yields superior results in both settings.

\textbf{Selection of Different CoT Constraints.}
We then analyze the impact of different CoT constraint strategies. While both strategies utilize the same three-stage reasoning process (Scene Description, Degradation Identification, and Restoration Plan), they differ in supervision. The `Union' strategy (d) treats the outputs of these stages as a single holistic sequence for supervision. In contrast, the `Split' strategy (f) employs three dedicated heads to output the results of each stage separately, applying independent and fine-grained supervision to each. Comparing (c) (no constraint), (d), and (f), it is evident that introducing explicit structured CoT constraints consistently improves performance. While the Union strategy (d) shows marginal gains or even drops in Vague settings compared to (c), the Split strategy (f) outperforms both by a notable margin (e.g., 23.73 dB vs 22.01 dB on Vague and 24.36 dB vs 22.57 dB on Precise). This demonstrates that disentangling the reasoning process into structured steps enables the model to learn more interpretable and accurate intermediate representations, ultimately leading to better restoration mappings.

\begin{figure}[t]
    \centering
    \includegraphics[width=1\linewidth]{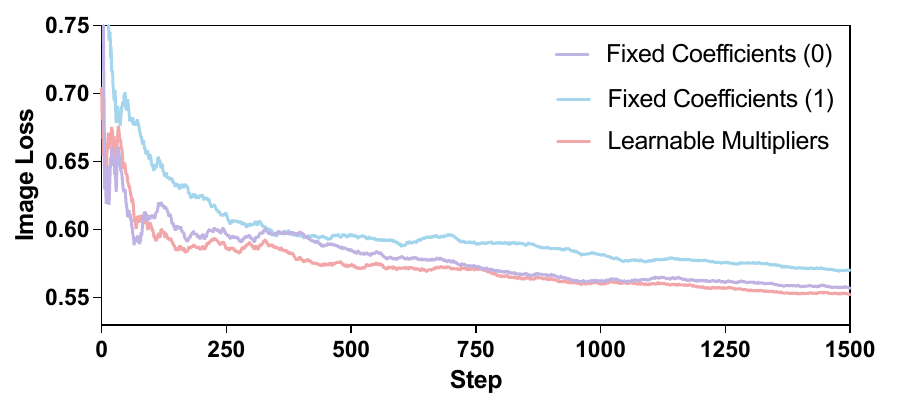}\\
    \vspace{-4mm}
    \caption{Training dynamics of restoration objective under three settings: Fixed Coefficient $1$, Fixed Coefficient $0$ (without CoT constraints), and Learnable Lagrange Multipliers.}
    \label{fig:lagrange_analysis}
    \vspace{-6mm}
\end{figure}

\textbf{Effectiveness of Lagrange Multipliers.}
We further examine the influence of the weighting strategy for the CoT loss components. We compare (e) Fixed Coefficients, where loss weights are static and the CoT loss is treated as a conventional penalty term, against (f) Learnable Coefficients, which employ Lagrange multipliers to dynamically adjust weights during training by treating them as dual variables jointly optimized with the model. As shown in Table~\ref{tab:ab_cot}, the learnable approach achieves significantly higher PSNR and lower LPIPS scores.
From an optimization perspective, updating Lagrange multipliers increases the weight of a CoT component when its corresponding constraint is violated and relaxes it once satisfied. This adaptive reweighting reduces sensitivity to coefficient tuning and prevents either the restoration objective or a single CoT component (scene description, degradation identification, and plan formulation) from dominating the gradients, leading to more stable convergence and better final performance.

\subsection{Analysis of Learnable Lagrange Multipliers} 
\label{sec:theory_lar}
This subsection presents an empirical comparison between learnable Lagrange multipliers and fixed coefficients, and further analyzes why the former is more compatible with optimizing the main restoration objective. As shown in Fig.~\ref{fig:lagrange_analysis}, during the early stage of training, the fixed-coefficient-1 setting incurs a substantially higher restoration loss, whereas the fixed-coefficient-0 and learnable-multiplier settings remain much closer, with the learnable setting consistently lower. This empirical observation suggests that overly strong fixed penalties can interfere with optimization of the restoration objective, while adaptive weighting mitigates such interference. 

To further analyze this results, we start from the training objective in Sec.~\ref{sec:problem}. Let $f(\theta)=\mathcal{L}_{\text{FM}}(\theta)$ and let $\psi_i(\theta)=\lVert \hat{c}_{i}(\theta)-c_i\rVert_2^2-\delta_i$ denote the violation of the $i$-th CoT constraint, where $i\in\{\text{s}, \text{d}, \text{p}\}$.

\begin{table*}[t]
\caption{Comparison of model parameter and runtime on 512$\times$512 images.}
\label{tab:runtime}
\centering
\vspace{-2mm}
\setlength{\tabcolsep}{1pt}
\resizebox{\linewidth}{!}{
\begin{tabular}{l|cccccccccccc}
\toprule
\textbf{Methods} & \textbf{PromptIR} & \textbf{OneRestore} & \textbf{InstructIR} & \textbf{DA-CLIP} & \textbf{HYPIR} & \textbf{LucidFlux} & \textbf{AutoDIR} & \textbf{DiffUIR} & \textbf{UniRestore} & \textbf{CoTIR}(Flux.1-12B) & \textbf{CoTIR}(Flux.2-4B) & \textbf{CoTIR}(Flux.2-9B) \\
\midrule
Params (B) & 0.04 & 0.01 & 0.03 & 0.05 & 1.13 & 13.62 & 1.49  & 0.04 & 1.07 & 13.47 & 4.98 & 11.05 \\
Times (s) & 0.23 & 0.02 & 0.03 & 11.68 & 0.07 & 8.11 & 10.97 & 0.30 & 0.28 & 7.95 (20 step) & 1.84 (5 step) & 2.44 (5 step) \\
\bottomrule
\end{tabular}}
\vspace{-6mm}
\end{table*}

\textbf{Fixed-weight penalty as a surrogate.}
We first consider the conventional fixed-weight penalty formulation with coefficients $\rho_i>0$:
\[
F(\theta)=f(\theta)+\sum_{i\in\{\text{s}, \text{d}, \text{p}\}}\rho_i\,[\psi_i(\theta)]_{+},
\label{eq:fixed_penalty}
\]
where $[x]_+=\max(x,0)$.
Let $\tilde{\theta}$ be a fixed strictly feasible point with $\psi_i(\tilde{\theta})\le -\sigma$ for all $i$ and some $\sigma>0$.
Let $\theta_{\rho}$ be a minimizer of $F$, assuming it is attained. Since $F(\theta_{\rho})\le F(\tilde{\theta})=f(\tilde{\theta})$, we have
\begin{equation}
\sum_{i\in\{\text{s}, \text{d}, \text{p}\}}\rho_i\,[\psi_i(\theta_{\rho})]_+ \le f(\tilde{\theta})-f(\theta_{\rho}) \le f(\tilde{\theta})- f^\star,
\label{eq:penalty_violation_bound}
\end{equation}
where $f^\star$ is the global lower bound of $f$. In particular, letting $\rho_{\min}=\min_i \rho_i$, Eq.~\eqref{eq:penalty_violation_bound} implies
\[
\sum_{i\in\{\text{s},\text{d},\text{p}\}} [\psi_i(\theta_\rho)]_+
\le \frac{f(\tilde{\theta})-f^\star}{\rho_{\min}}.
\]
Thus, this upper bound tightens only when the penalty coefficients increase. In other words, reducing the constraint violation under a fixed-penalty formulation generally requires sufficiently large penalty coefficients. However, large fixed penalties also amplify the gradients of active constraint terms, which can dominate the restoration objective and slow down optimization. This explains why the fixed-coefficient-1 setting in Fig.~\ref{fig:lagrange_analysis} exhibits the highest restoration loss in the early stage, and why the fixed-coefficient setting in Table~\ref{tab:ab_cot} (e) underperforms the learnable-multiplier setting (f).

\textbf{Learnable multipliers under dual-optimizer training.}
Instead of enforcing all constraints with fixed strength throughout training, we introduce learnable multipliers through the Lagrangian
\[
\mathcal{L}(\theta,\lambda)= f(\theta)+\sum_{i\in\{\text{s}, \text{d}, \text{p}\}}\lambda_i\, \psi_i(\theta),\quad \lambda\ge 0,
\]
where $\lambda=(\lambda_s, \lambda_d, \lambda_p)^\top$. In practice, we optimize the model parameters and the multipliers jointly with two optimizers, while enforcing non-negativity on the multipliers. Each multiplier is adjusted according to the violation of its corresponding CoT constraint: larger violation leads to a larger multiplier, whereas satisfied constraints keep the multiplier small. Compared with fixed penalties, this mechanism keeps the optimization close to that without CoT constraint terms, thereby achieving a more favorable balance and optimal restoration performance.

\begin{figure}[t]
    \centering
    \includegraphics[width=1\linewidth]{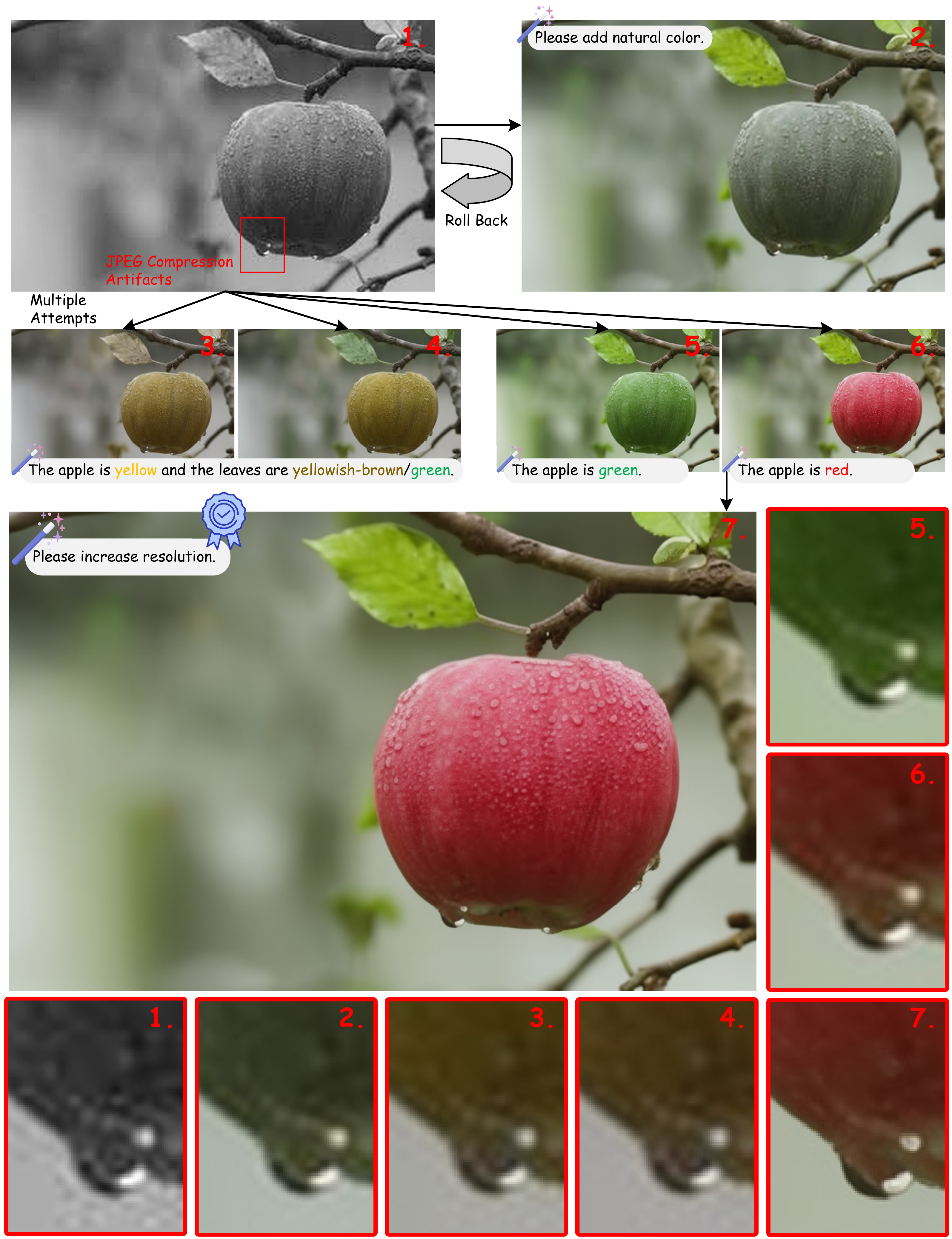}
    \vspace{-6mm}
    \caption{Case of multi-turn restoration. The proposed method accurately follows commands to complete the restoration.}
    \vspace{-5mm}
    \label{fig:results_use_case_multi}
\end{figure}

\subsection{Application of Multi-round Restoration}
Although the proposed CoTIR is designed to support autonomous restoration through internalized reasoning, restoration models that can follow user intent are often of greater practical value in real applications. Fig.~\ref{fig:results_use_case_multi} shows a subjective multi-turn restoration case containing JPEG compression artifacts and color fading. In this example, the result obtained by direct automatic restoration is not sufficiently satisfactory to the user. The user then provides additional color preferences and finally requests resolution enhancement. Throughout this multi-turn process, CoTIR accurately follows the user instructions and progressively produces the desired result. This example demonstrates that the proposed method also supports fine-grained multi-turn restoration for meeting diverse user needs.

\subsection{Efficiency Comparison}
Table~\ref{tab:runtime} compares the model size and inference time on 512$\times$512 images. Conventional CNN/Transformer-based methods are generally lightweight and fast, but their restoration capacity is limited compared with recent large generative models. Among the latter, CoTIR achieves a more favorable efficiency-performance trade-off by internalizing reasoning within a single restoration model instead of relying on sequential tool invocation or long multi-stage processing. In particular, CoTIR (Flux.2-4B) requires only 1.84s with 5 sampling steps, which is substantially faster than DA-CLIP, LucidFlux, and AutoDIR, while still maintaining strong restoration performance. Although the Flux.1-12B and Flux.2-9B variants have larger model sizes, they remain competitive in runtime and provide scalable choices for balancing restoration quality and efficiency.

\section{Conclusion}

In this paper, we have presented CoTIR, a universal image restoration framework that internalizes Chain-of-Thought (CoT) reasoning into a single model. Rather than explicitly chaining multiple restoration stages, CoTIR encodes structured reasoning as differentiable constraints in a Lagrangian-inspired objective, enabling holistic correction of complex, mixed degradations while better modeling their interactions. Additionally, we have contributed CoTIR-Bench, a large-scale benchmark comprising 5.2 million samples annotated with high-quality CoT reasoning data for each pair. Extensive experiments have demonstrated that CoTIR achieves SOTA performance across diverse restoration scenarios.

% \textbf{Limitations.} Our CoTIR targets high-quality restoration for complex degradations and requires a relatively high computational budget compared with traditional CNN- or Transformer-based methods. Additionally, in severely degraded scenarios, the inherent generative prior and CoT reasoning may introduce spurious details or remove subtle objects. Future work will focus on improving efficiency and robustness in such extreme cases. 

% or


%\input{sections/supplement.tex}

% % you can choose not to have a title for an appendix
% % if you want by leaving the argument blank
% \section{}

% \input{sectkyions/acknowledgement.tex}

\ifCLASSOPTIONcaptionsoff
  \newpage
\fi



%\bibliographystyle{IEEEtran}
%\begin{thebibliography}{1}
% \newpage
\bibliographystyle{IEEEtran}
\bibliography{references}
%\end{thebibliography}
% \newpage
% \input{sections/supplement.tex}

% \newpage

% \begin{IEEEbiography}
% [{\includegraphics[width=1in,height=1.25in,clip,keepaspectratio]{./portraits/YuanGao.jpg}}]{Yuan Gao} is currently pursuing the Ph.D. degree in Tsinghua University, Beijing, China. His research interests include Scientific Machine Learning and Computer Vision. He has published papers on top-tier conferences and journals such as ICML, AAAI, ECCV, IEEE TITS, IEEE TIM, and KBS.
% \end{IEEEbiography}

% \clearpage
% \appendix
% \input{sections/supplement.tex}

\end{document}